\documentclass[sn-apa]{sn-jnl}% APA Reference Style
\jyear{2022}%
\usepackage{subcaption}

\theoremstyle{thmstyleone}%
\theoremstyle{thmstyletwo}%

\theoremstyle{thmstylethree}%

\usepackage{amsmath}
\usepackage{graphicx}
\newsavebox\CBox
\def\textBF#1{\sbox\CBox{#1}\resizebox{\wd\CBox}{\ht\CBox}{\textbf{#1}}}

\begin{document}

\title[FES for CDFSL]{Feature Extractor Stacking for Cross-domain Few-shot Learning}

\author*{\fnm{Hongyu} \sur{Wang}} \email{hw168@students.waikato.ac.nz}

\author{\fnm{Eibe} \sur{Frank}}% \email{eibe@waikato.ac.nz}

\author{\fnm{Bernhard} \sur{Pfahringer}} %\email{bernhard@waikato.ac.nz}

\author{\fnm{Michael} \sur{Mayo}}% \email{mmayo@waikato.ac.nz}

\author{\fnm{Geoffrey} \sur{Holmes}} %\email{geoff@waikato.ac.nz}

\affil{\orgdiv{Department of Computer Science}, \orgname{University of Waikato}, \orgaddress{\street{Knighton Road}, \city{Hamilton}, \postcode{3240}, \state{Waikato}, \country{New Zealand}}}

\abstract{Cross-domain few-shot learning (CDFSL) addresses learning problems where knowledge needs to be transferred from one or more source domains into an instance-scarce target domain with an explicitly different distribution. Recently published CDFSL methods generally construct a universal model that combines knowledge of multiple source domains into one feature extractor. This enables efficient inference but necessitates re-computation of the extractor whenever a new source domain is added. Some of these methods are also incompatible with heterogeneous source domain extractor architectures. We propose feature extractor stacking (FES), a new CDFSL method for combining information from a collection of extractors, that can utilise heterogeneous pretrained extractors out of the box and does not maintain a universal model that needs to be re-computed when its extractor collection is updated. We present the basic FES algorithm, which is inspired by the classic stacked generalisation approach, and also introduce two variants: convolutional FES (ConFES) and regularised FES (ReFES). Given a target-domain task, these algorithms fine-tune each extractor independently, use cross-validation to extract training data for stacked generalisation from the support set, and learn a simple linear stacking classifier from this data. We evaluate our FES methods on the well-known Meta-Dataset benchmark, targeting image classification with convolutional neural networks, and show that they can achieve state-of-the-art performance.}


\maketitle

\section{Introduction}

Cross-domain few-shot learning (CDFSL) addresses the problem that deep learning methods, such as convolutional neural networks (CNN) for image classification, generally require a large amount of labelled training data to achieve high predictive accuracy when trained from scratch. CDFSL algorithms are designed for scenarios where only a few labelled training instances are available in the form of a so-called ``support set". The aim is to nevertheless achieve high accuracy when predicting labels for instances of the target domain that have never been seen before, i.e., the so-called ``query set". This can generally only be achieved by applying transfer learning: taking knowledge gleaned from one or several source domains with large-scale training data and using this knowledge to inform learning in a few-shot target domain.

In CDFSL, the source domain(s) and the target domain are assumed to have potentially very distinct properties. This cross-domain setting is arguably more realistic than the ``in-domain" scenario, used in some few-shot learning literature~\citep{vinyals2016matching}, where the source and the target domains comprise mutually exclusive sets of classes obtained from the same dataset.\footnote{Note that,  strictly speaking, this also creates distinct domains because the joint probability distributions will differ. However, they will be strongly related.} It also yields harder learning problems due to greater domain shift. 

In a CDFSL setting with multiple source domains, an algorithm needs to both select relevant source domains and effectively transfer their knowledge into a target domain using a few-shot support set. Performance is measured by ``meta-testing"---transferring model(s) using target domain support sets and evaluating their predictive accuracy on corresponding query sets. Recent work considering performance in image classification, which is the setting we also focus on in this paper, shows that single-domain learning (SDL) and vanilla multi-domain learning (MDL), which applies one feature extractor and multiple classification heads, fail to achieve competitive performance compared to methods specifically designed for CDFSL~\citep{Meta-Dataset2020Triantafillou,li2021universal}.

A majority of recently published CDFSL methods involve building a universal model from a collection of extractors, with each extractor pretrained in a distinct source domain. This comprises the so-called ``meta-training" phase, which is performed before meta-testing begins. The universal-model paradigm is generally efficient when performing meta-testing because a single universal feature extractor is used and fine-tuned on the support set, usually in conjunction with a simple robust classifier that turns extracted feature vectors into predictions. However, training the universal model is computationally expensive, and some methods constrain all extractors to the same architecture as the intended universal model~\citep{triantafillou2021learning}, rendering them inapplicable to heterogeneous extractor collections that are likely to occur in real-world practice. Moreover, they may require adjustment based on pre-existing domain knowledge to function well. For example, given a source domain/extractor collection for image classification consisting of ImageNet~\citep{deng2009imagenet, russakovsky2015imagenet}, along with other, less comprehensive source domains, authors often assign greater importance to the ImageNet extractor during training~\citep{triantafillou2021learning, li2021universal}. This achieves good performance on benchmarks, which normally include target domains such as CIFAR-10 that are quite similar to ImageNet in nature, but may not be as useful in real-world applications involving less similar data. Lastly, the process of deriving a universal model is non-incremental, which means it needs to be re-run whenever an extractor is updated or added, and normally requires access to the entire meta-training dataset~\citep{triantafillou2021learning, li2021universal}.

As an alternative approach that avoids these shortcomings, we propose a novel ``lazy" CDFSL method, termed feature extractor stacking (FES), that fine-tunes each extractor independently and trains a classifier using a form of stacked generalisation~\citep{wolpert1992stacked} during meta-testing. The ``meta-training" phase in FES consists solely of training individual feature extractors, one for each source domain, using standard single-domain supervised learning. In practical applications, it may be possible to skip meta-training entirely if a set of suitable feature extractors has been obtained from other sources. FES is fully compatible with heterogeneous extractor collections, imposing no constraints on their architecture or fine-tuning configuration. It assumes equal importance of the extractors {\em a priori}, determining their task-specific relevance based purely on the support set, and does not require derivation of a universal model.

Along with the basic FES algorithm, which applies a simple linear stacking classifier and is described in Section \ref{sec:FES}, we present two variants: convolutional FES (ConFES) in Section \ref{sec:ConFES} and regularised FES (ReFES) in Section \ref{sec:ReFES}. ConFES replaces the flat global kernel of FES with a hierarchy of depthwise convolutional kernels, reducing the number of parameters in the stacking classifier. ReFES applies fused lasso regularisation~\citep{tibshirani2005sparsity} to the stacking classifier of FES to reduce the weights of irrelevant snapshots and induce smooth weight transition between adjacent snapshots.

We evaluate FES and its variants on the Meta-Dataset benchmark~\citep{Meta-Dataset2020Triantafillou}, which contains eight source domains and five target domains, and include five additional target domains: CropDisease, EuroSAT, ISIC, ChestX, and Food101~\citep{guo2020broader, bossard2014food}. We show that FES outperforms three recent universal-model methods: URL~\citep{li2021universal}, FLUTE~\citep{triantafillou2021learning}, and a URL extractor with TSA fine-tuning~\citep{li2022cross}, and advances the state of the art on Meta-Dataset. We also discuss practical advantages of FES in real-world scenarios, as FES can work with heterogeneous extractors out of the box and does not need to train a universal model.

\section{Related work}

Our empirical comparison of CDFSL methods is based on the Meta-Dataset framework, so we review this benchmark first before discussing methods that we compare to our approach. We also briefly review other noteworthy methods in the literature.

\subsection{The Meta-Dataset benchmark}\label{sec:meta-dataset}

The Meta-Dataset~\citep{Meta-Dataset2020Triantafillou} benchmark has multiple configurations; we describe the CDFSL configuration that we use---most recent publications in the field use this configuration as well. It contains eight source domains: ILSVRC-2012 (ImageNet), Omniglot, Aircraft, CUB-200-2011 (Birds), Describable Textures, Quick Draw, Fungi, and VGG Flower. Recent work utilising Meta-Dataset~\citep{requeima2019fast} has extended its original set of two target domains, Traffic Signs and MSCOCO, by adding three more: MNIST, CIFAR10, and CIFAR100. For an even more comprehensive evaluation, we add four target domains from the CDFSL benchmark in~\citet{guo2020broader}---CropDisease, EuroSAT, ISIC, and ChestX---but additionally also employ Food101~\citep{bossard2014food}. Only the 250 sanitised test images in each Food101 class are used in our experiments.

The Meta-Dataset framework splits each source domain into three partitions: training, validation, and test. The partitions are mutually exclusive in terms of their classes, with the training partition containing approximately 70\% of source domain classes and the validation and test partitions containing approximately 15\% each. The training and validation partitions are made available to the CDFSL method for ``meta-training", where the training partition is generally used to train extractors and the validation split to aid hyperparameter tuning. The test partition is reserved for evaluating the CDFSL method by sampling few-shot episodes (i.e., meta-testing): the term ``episode'' refers to the process of sampling a support set and a query set, training a classifier on the support set, and evaluating it on the query set.

In contrast, the entire target domain data can be used for sampling episodes to evaluate few-shot learning in these domains. It is important to note that, by definition, only tasks sampled from target domains truly measure CDFSL performance. Using terminology that is common in this context, good performance in these domains indicates ``strong generalisation"; good performance on tasks sampled from source domain test partitions indicates ``weak generalisation".

The most commonly used method to evaluate CDFSL algorithms on Meta-Dataset is to generate 600 any-way any-shot episodes from each dataset, and measure each algorithm's mean classification accuracy on these 600 episodes, as well as the 95\% confidence interval. Any-way any-shot sampling means the number of classes for each episode and the number of support instances per class are arbitrary, leading to imbalanced support sets more representative of real-world scenarios than fixed-way fixed-shot episodes. The query set is balanced in the Meta-Dataset setting. We adhere to this evaluation method in our experiments.

\subsection{Methods included in the experimental comparison}

Two recently published CDFSL methods that advanced the state-of-the-art on Meta-Dataset are Few-shot Learning with a Universal TEmplate (FLUTE)~\citep{triantafillou2021learning} and Universal Representation Learning (URL)~\citep{li2021universal}. Even more recently, based on a URL universal model, a fine-tuning method using Task-Specific Adaptors (TSA)~\citep{li2022cross} improved accuracy on some target domains even further. We compare our new FES approach to these methods in our experiments.

\subsubsection{Few-shot Learning with a Universal TEmplate.}

Based on the FiLM approach~\citep{perez2018film}, FLUTE trains a universal model in the source domains, employing the ResNet18 architecture~\citep{he2016deep} widely used in CDFSL, but maintaining a separate set of batch normalisation~\citep{ioffe2015batch} parameters for each domain. The ResNet ``template" contains one set of convolutional weights shared across all source domains, and only the batch normalisation parameters are specific to each source domain. FLUTE jointly trains the template in all source domains. At each training iteration, a random source domain is selected---with ImageNet having a 50\% probability of being selected and the other seven source domains evenly sharing the other 50\% probability---and a batch of input data is sampled from the selected source domain. In forward propagation, the input batch flows through the shared convolutional layers and the selected domain's set of batch normalisation layers, and loss is computed by applying a cosine classifier~\citep{chen2019closer, chen2021meta}. A nuance of FLUTE training is that backpropagation is performed using a ``meta-batch" of eight individual batches: the intention is to stabilise training by aggregating loss values across multiple domains. Hyperparameter tuning is performed using episodes sampled from source domain validation partitions.

When the template is trained, snapshots are frequently saved. The final template is chosen as the snapshot that performs best on the source domains' validation partitions. To establish performance, few-shot episodes are sampled from these partitions. For each episode, feature vectors are extracted using the shared convolutional layers and the domain's set of batch normalisation layers. Accuracy is computed using a nearest-centroid classifier~\citep{mensink2013distance, snell2017prototypical}.

One more component of FLUTE, produced in a separate meta-training phase, is a blender network, which is a dataset classifier based on a permutation-invariant set encoder~\citep{zaheer2017deep} followed by a linear layer. Given a batch of instances, the blender predicts, as a probability distribution, the source domain from which the batch is sampled. It is trained on batches sampled from the source domains' training partitions, and the final blender model is chosen using batches from the validation partitions.

Given a few-shot episode at meta-test time, the blender uses the support set to produce a probability distribution. These probabilities in turn are used to form a linear combination of the source-domain-specific batch normalisation weights. Along with the shared convolutional weights from the template, this forms the initial set of parameters for the ResNet18 feature extractor, which is applied in conjunction with a nearest-centroid classifier. The model's batch normalisation parameters are then fine-tuned on the support set while its convolutional weights remain fixed.

\subsubsection{Universal Representation Learning}

The URL algorithm also generates a universal model. It first pretrains domain-specific ResNet18 extractors independently. Then, a separate ResNet18 feature extractor is trained to form a universal model by distillation. This model is trained to match each extractor's output feature vectors and logits using instances sampled from the extractor's corresponding domain. To this end, the universal model contains pairs of auxiliary domain-specific components that each comprise 1) a projection layer that transforms the universal extractor's feature vectors to match those of each domain-specific extractor, and 2) a classifier layer trained to match the logits produced by each extractor. 

In the experiments by~\citet{li2021universal}, ImageNet is made more prominent in distillation: ImageNet instances make up 50\% of each mini-batch and the other seven source domains evenly make up the rest. Snapshots of the universal feature extractor are saved at predefined intervals during knowledge distillation. Episodes sampled from source domain validation partitions are used to select the best snapshot as a form of early stopping.

After meta-training, the auxiliary components of the universal model are discarded, leaving only the feature extractor. During meta-testing, this extractor is frozen, and a projection layer is initialised with an identity weight matrix and trained using the support set. The projected feature vectors are used to build a nearest-centroid classifier. Cosine similarity values between a feature vector to be classified and the centroids are used as logits. Fine-tuning minimises cross-entropy loss on the support set. Note that during fine-tuning, as the projection layer is optimised, projected support feature vectors change, and their centroids change as well. The fine-tuning effect can be interpreted as forming better clusters with projected support feature vectors.

\subsubsection{Task-Specific Adaptors}

TSA~\citep{li2022cross} is a fine-tuning method suitable for CDFSL. Given a pretrained extractor, trainable task-specific adaptors are attached to it, and the support set is used to optimise the adaptors with the extractor's original weights frozen. Like URL, TSA also attaches a trainable linear projection layer and a robust classifier to the end of the feature extractor during fine-tuning, but it adds further adaptor components. Among multiple configurations examined, the most effective approach for few-shot image classification found by~\citet{li2022cross} is to attach channel projection matrices as residual connections to a model's convolutional layers. \citet{li2022cross} used TSA in conjunction with a URL-distilled universal extractor, but TSA can be applied to other CNN architectures as well.

\subsection{Other work on CDFSL}

We review additional noteworthy CDFSL methods here. These methods precede FLUTE, URL, and TSA chronologically and achieve lower accuracy than results presented by~\citet{triantafillou2021learning} and ~\citet{li2021universal,li2022cross}. Hence, in the experiments presented in this paper, we only compare to FLUTE, URL, and a URL extractor with TSA fine-tuning. 

\subsubsection{Selecting Relevant Features from a Universal Representation}

SUR~\citep{dvornik2020selecting} is a CDFSL method that utilises independently pretrained feature extractors directly for meta-testing. Each extractor is used to extract a set of feature vectors from the support set, with a trainable weight assigned to it. Feature vectors are multiplied by their respective weights and concatenated to provide input to a nearest-centroid classifier. The weights are trained by optimising loss of the classifier on the support set. SUR is similar to URL in the meta-testing phase, as both make predictions with a nearest-centroid classifier and optimise parameters on the support set; the primary difference is that URL maintains a universal model while SUR uses the original extractors directly.

\subsubsection{Universal Representation Transformer}

URT~\citep{liu2020universal} also assigns a weight to each source domain extractor during meta-testing. However, it utilises a weight assignment model learned using meta-training instead of direct optimisation on the support set to obtain the weights. To this end, URT trains an attention mechanism~\citep{vaswani2017attention} that learns to assign appropriate weights to source domain feature extractors given a few-shot episode. The weight assignment model is trained and has its hyperparameters selected using episodes sampled from the source domains' training and validation partitions.

\subsubsection{Conditional Neural Adaptive Processes}

The CNAPs method, as proposed in ~\citet{requeima2019fast}, uses an extractor pretrained in a large source domain, e.g., ImageNet~\citep{deng2009imagenet, russakovsky2015imagenet}, and meta-trains adaptation networks, using episodes sampled from the source domains, to produce task-specific FiLM~\citep{perez2018film} transformations and a linear classifier for each few-shot episode.

A variant, Simple CNAPs~\citep{bateni2020improved}, was later proposed utilising a non-parametric Mahalanobis distance~\citep{galeano2015mahalanobis} measure in place of the classifier adaptation network of CNAPs, reducing the parameter count and improving CDFSL performance. A transductive version of Simple CNAPs was subsequently also proposed~\citep{bateni2022enhancing}, making use of clustering of query instances in feature space to achieve better performance than Simple CNAPs, assuming that the query set is available as a batch instead of a sequential stream of incoming instances. As most other CDFSL methods do not rely on such an assumption, they cannot be compared to transductive CNAPs on an even footing.

\subsubsection{Multi-Mode Modulator}

Tri-M~\citep{liu2021multi}, akin to CNAPs, uses an extractor pretrained in a large-scale source domain, and meta-trains a modulation network using source domain episodes to generate appropriate FiLM transformations for each few-shot episode. Tri-M maintains two sets of transformations---a domain-specific one and a domain-cooperative one---and its resulting FiLM transformation is a combination of the two. Tri-M determines a source domain for its domain-specific transformation in a way similar to how FLUTE~\citep{triantafillou2021learning} utilises its blender network and uses an attention mechanism~\citep{vaswani2017attention} to compose its domain-cooperative transformation from relevant source domains.

\section{Cross-domain Few-shot Learning using Stacking}

Considering the CDFSL methods discussed in the previous section, the SUR method stands out because its meta-training process is straightforward: all it involves is pretraining individual source domain feature extractors. Once these have been obtained, SUR performs ``lazy'' learning in the sense that significant work is only performed once the support set for a few-shot episode becomes available. This makes it very flexible because new extractors can be added at any time. However, SUR does not yield state-of-the-art performance. The new methods presented in this paper are inspired by SUR and the old and established method of applying stacked generalisation to learning a classifier that combines predictions of multiple base classifiers. Henceforth, we will refer to this classifier as the ``stacking classifier". There are four primary differences between SUR and our stacking-based methods: 1) the source domain extractors are fined-tuned on the support set to extract more information from this data by attaching appropriate classifier layers to them, 2) two-fold cross-validation is used to generate training data for the stacking classifier to tackle overfitting, 3) the feature vectors of this training data consist of logits obtained from classifier layers attached to the extractors, and 4) multiple snapshots of each extractor are stored during fine-tuning and used to obtain sets of logits, adding further richness to the data available for training the stacking classifier.

In the following, we first explain the basic method of feature extractor stacking (FES) in detail and prove convexity of its optimisation, before describing two variants: convolutional FES (ConFES) and regularised FES (ReFES).

\subsection{Feature Extractor Stacking}\label{sec:FES}

\begin{sidewaysfigure}
    \includegraphics[width=\textwidth]{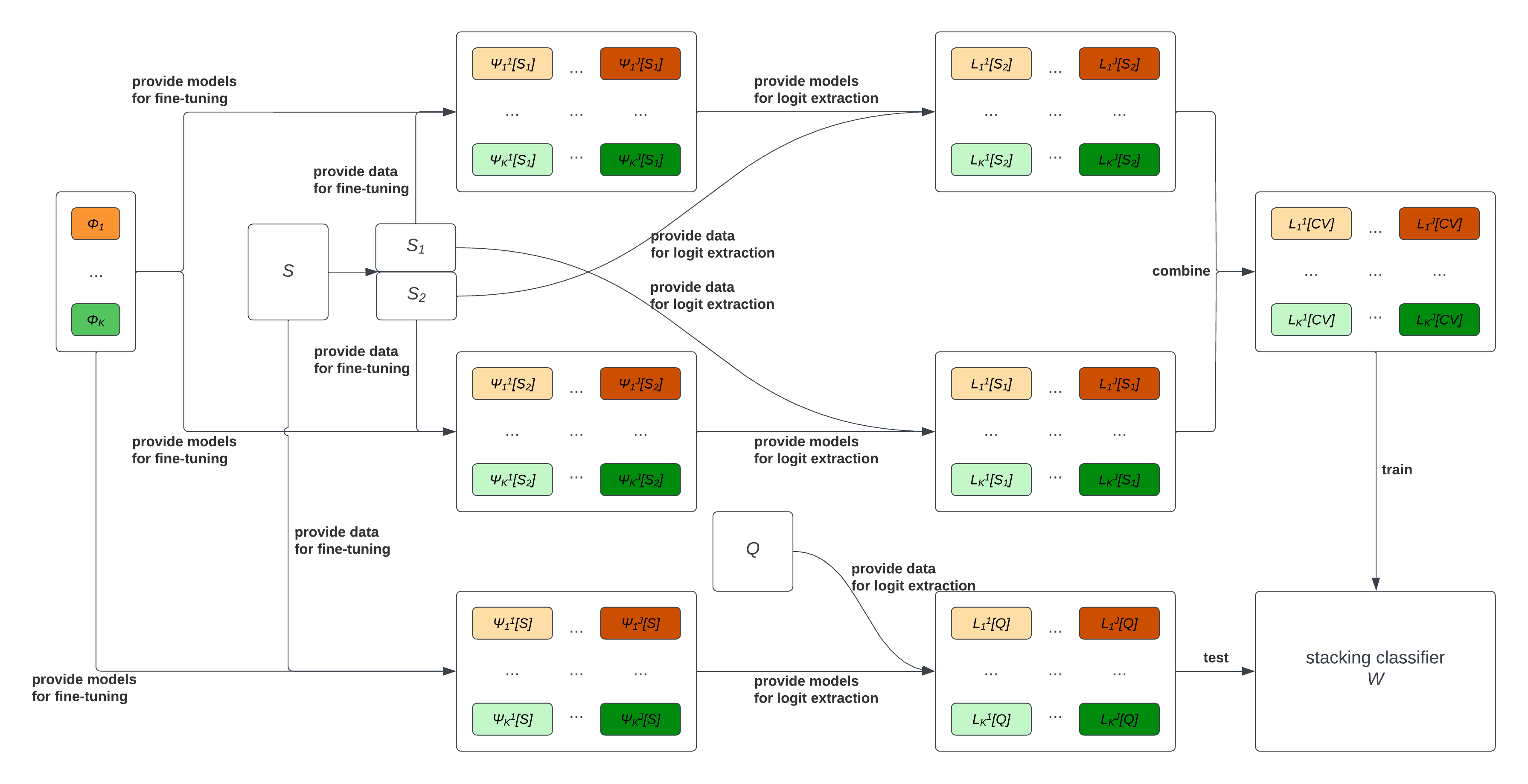}
    \caption{Framework of FES. Given an extractor collection with $K$ extractors, each extractor $\Phi$ is set up as a network $\Psi$ for fine-tuning. The support set $S$ is split into $S_1$ and $S_2$ using stratified cross-validation. Each network $\Psi$ is fine-tuned on one split, producing $J$ snapshots in the process, and these snapshots are used to extract logits from the other split. Logits extracted from both splits are combined into cross-validated logits of the full support set, which are used to train a stacking classifier $W$ to fit $S$'s labels. The full $S$ is then used to fine-tune $\Psi$, producing snapshots to extract logits for the query set $Q$. $W$ takes $Q$'s logits as input and predicts $Q$'s labels.}
    \label{fig:FES_framework}
\end{sidewaysfigure}

Given pretrained feature extractors, FES has three key components: fine-tuning extractors to obtain snapshots, two-fold cross-validation to produce training data for the stacking classifier, and training of the stacking classifier. Figure \ref{fig:FES_framework} depicts the FES framework.

\subsubsection{Fine-tuning the extractors}

We use $f_{\Phi_1}, f_{\Phi_2}, ..., f_{\Phi_{K}}$ (or just $\Phi_1, \Phi_2, ..., \Phi_{K}$ for brevity) to denote the collection of pretrained feature extractors, where $\Phi$ represents the corresponding extractor's parameters and $K$ is the number of source domains. The support set of a few-shot episode is denoted $S$ and the query set $Q$. $S$ contains $N$ instances belonging to $C$ classes. We fine-tune each extractor independently on $S$. As $f_{\Phi}$ is a feature extractor, a classifier $g$ with parameters $\Theta_1$ is attached to $f_{\Phi}$ to produce logits. Auxiliary components with parameters $\Theta_2$ may also be introduced to the model to aid fine-tuning, such as with TSA~\citep{li2022cross}. The resulting model is defined as $h_{\Psi} = g_{\Theta_1} \circ f_{(\Phi, \Theta_2)}$, where we use $\Psi$ to denote the combination of all parameters. It is possible for $\Theta_2$ to be $\varnothing$, as auxiliary fine-tuning components are optional. $J$ snapshots are saved sequentially at different fine-tuning iterations of $h_{\Psi}$. Each snapshot contains parameters $\Psi_k^j[S]$, where $k \in [1,K]$ and $j \in [1,J]$, with $S$ denoting the fine-tuning set used.

\subsubsection{Cross-validation to obtain training data for stacked generalisation}

In stacked generalisation~\citep{wolpert1992stacked}, cross-validation is employed to obtain training data for the stacking classifier to combat overfitting, and it is applied in FES as well. More specifically, we apply stratified two-fold cross-validation to the support set $S$, producing two splits $S_1$ and $S_2$, which will take turns serving as the training split $S^{train}$ and the test split $S^{test}$. It is possible to employ more folds in FES, but using additional folds did not yield performance gains in our experiments. 

Training on one of the training splits amounts to fine-tuning a network $h_{\Psi}$ on this data. In principle, this could be done for a fixed number of iterations, and once complete, logits on the corresponding test split could be obtained as training data for the stacking classifier. However, this naive approach may not work well because it is not known how many iterations should be performed for fine-tuning to maximise accuracy of the full learning system. The approach we propose and evaluate in this paper is instead based on the idea that we can take multiple snapshots of the models during fine-tuning and use all the snapshots' logits on the test folds for training the stacking classifier. In other words, the learning algorithm for the stacking classifier will be responsible for deciding which extractor snapshots are the most useful ones for making accurate predictions on the test folds.

More specifically, given a pair $(S^{train}, S^{test})$ and an extractor $h_{\Psi}$, we fine-tune $h_{\Psi}$ on $S^{train}$ with the same configuration used to obtain $h_{\Psi^j[S]}$, e.g., optimiser, learning rate, etc., and save snapshots $h_{\Psi^j[S^{train}]}$ at the same iterations as $h_{\Psi^j[S]}$. Logits $L^j[S^{test}]$ are extracted from $S^{test}$ with each $h_{\Psi^j[S^{train}]}$, i.e., $L^j[S^{test}] = h_{\Psi^j[S^{train}]}(S^{test})$. Using this approach, the two splits $S_1$ and $S_2$ can be used to alternately fine-tune extractors and produce logits $L^j[S_1]$ and $L^j[S_2]$, which are combined into $L^j[CV]$, i.e., logits for every support set instance extracted using cross-validation. Considering the logits from all $K$ extractors jointly, $L_K^J[CV]$ is a tensor of shape $N \times K \times J \times C$, i.e., $N$ support instances converted into logits for $C$ classes extracted by $K \times J$ snapshot models, ready to serve as training data for the stacking classifier.

\subsubsection{Stacking classifier training}

\begin{figure}[t]
    \centering
    \includegraphics[width=\textwidth]{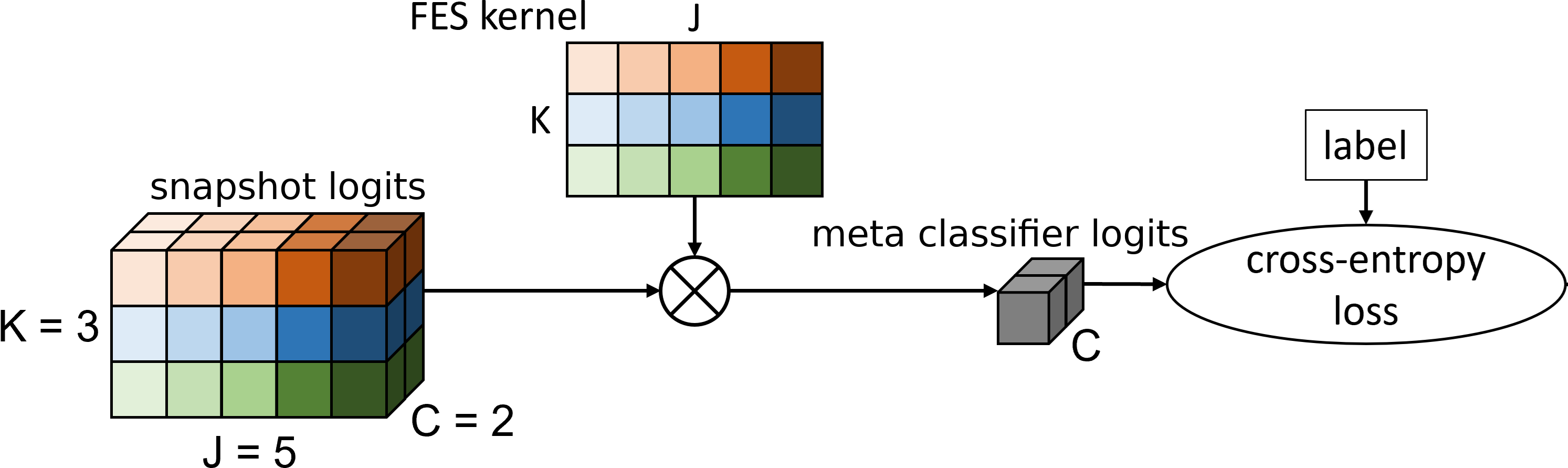}
    \caption{FES uses a global kernel to compute stacking classifier logits from the snapshots' base logits. The global kernel is essentially flat since it makes no use of the snapshots' temporal relations. For demonstration purposes, this figure and the following ones assume three extractors ($K = 3$), five fine-tuning snapshots per extractor ($J = 5$), and a two-class problem ($C = 2$).}
    \label{fig:FES_metaclassifier}
\end{figure}

The FES stacking classifier is a weight matrix $W$ of shape $K \times J$, with $W_k^j$ representing $\Psi_k^j$'s weight. Given an instance $l$ of shape $K \times J \times C$, the stacking classifier's output logits $l^W$ are obtained using a simple weighted average:
\begin{equation}\label{eq:w}
    l^W[c] = \sum_{k = 1}^K\sum_{j = 1}^JW_k^j \cdot l_k^j[c],
\end{equation}
where $c$ is one of the $C$ classes. We compute the cross-entropy loss using the $N$ support set logits $L^W$ output by the stacking classifier and the one-hot-encoded labels $Y$, i.e., $-\sum\limits_{n = 1}^NY_n \log(\text{softmax}(L_n^W))$, which we minimise by training $W$. For interpretability, we constrain all values in $W$ to be non-negative by clipping negative weights with ReLU. The FES stacking classifier is shown in Figure \ref{fig:FES_metaclassifier}.

After training, $W$ is used with Equation \ref{eq:w} to compute meta logits for the query set $Q$ using the logits $L_K^J[Q]$ computed by the saved snapshots $\Psi_K^J[S]$. Then, a softmax function is used to obtain class probability estimates.

\subsection{Proof of Convexity}

Given a stacking instance $l$ consisting of base logits obtained from the extractor snapshots, which the stacking classifier transforms into meta-level logits $l^W$, and the label $c_y$, the negative log-likelihood loss $\ell$ associated with the stacking classifier's parameters $W$ is
\begin{equation}\label{eq:loss}
    \ell(W) = \log(\sum_{i = 1}^Ce^{l^W[c_i]}) - l^W[c_y].
\end{equation}
To prove that optimising FES is a convex problem, we show that for any two values of $W$, named $A$ and $B$, a linear combination of the loss on $A$ and the loss on $B$ is never smaller than the loss obtained for the corresponding linear combination of the parameter values $A$ and $B$, i.e.,
\begin{equation}\label{eq:loss_inequality}
    \ell(\lambda A + (1 - \lambda) B) \leq \lambda\ell(A) + (1 - \lambda)\ell(B), \lambda \in [0, 1].
\end{equation}
Applying Equation \ref{eq:loss} to Equation \ref{eq:loss_inequality}, we get
\begin{align*}
    & \log(\sum_{i = 1}^Ce^{l^{(\lambda A + (1 - \lambda) B)}[c_i]}) - l^{(\lambda A + (1 - \lambda) B)}[c_y] \leq \\
    & \lambda(\log(\sum_{i = 1}^Ce^{l^A[c_i]}) - l^A[c_y]) + (1 - \lambda)(\log(\sum_{i = 1}^Ce^{l^B[c_i]}) - l^B[c_y]),
\end{align*}
which can be simplified into
\begin{equation}\label{eq:inequality_simplified}
    \log(\sum_{i = 1}^Ce^{l^{(\lambda A + (1 - \lambda) B)}[c_i]}) \leq \lambda\log(\sum_{i = 1}^Ce^{l^A[c_i]}) + (1 - \lambda)\log(\sum_{i = 1}^Ce^{l^B[c_i]}),
\end{equation}
because using Equation \ref{eq:w}, we have
\begin{align*}
    & l^{(\lambda A + (1 - \lambda) B)}[c_y] \\
    = & \sum_{k = 1}^K\sum_{j = 1}^J(\lambda A_k^j + (1 - \lambda) B_k^j) \cdot l_k^j[c_y] \\
    = & \sum_{k = 1}^K\sum_{j = 1}^J\lambda A_k^j \cdot l_k^j[c_y] + \sum_{k = 1}^K\sum_{j = 1}^J(1 - \lambda) B_k^j \cdot l_k^j[c_y] \\
    = & \lambda\sum_{k = 1}^K\sum_{j = 1}^JA_k^j \cdot l_k^j[c_y] + (1 - \lambda)\sum_{k = 1}^K\sum_{j = 1}^JB_k^j \cdot l_k^j[c_y] \\
    = & \lambda l^A[c_y] + (1 - \lambda) l^B[c_y].
\end{align*}
Similarly, Equation \ref{eq:inequality_simplified} can be transformed using Equation \ref{eq:w} into
\begin{equation}\label{eq:meta_logit_lse}
    \log(\sum_{i = 1}^Ce^{\lambda l^A[c_i] + (1 - \lambda) l^B[c_i]}) \leq \lambda\log(\sum_{i = 1}^Ce^{l^A[c_i]}) + (1 - \lambda)\log(\sum_{i = 1}^Ce^{l^B[c_i]}).
\end{equation}
It is known that the LogSumExp function $LSE(x) = \log(\sum\limits_{i = 1}^ne^{x_i})$ is convex. Therefore, we have
\begin{equation}\label{eq:standard_lse}
    \forall n \in \mathbb{Z}^+, \alpha, \beta \in \mathbb{R}^n : LSE(\lambda \alpha + (1 - \lambda) \beta) \leq \lambda LSE(\alpha) + (1 - \lambda) LSE(\beta).
\end{equation}
Hence, Equation \ref{eq:meta_logit_lse} is true because we can make the following assignments:
\begin{align*}
    n & = C, \\
    \alpha_i & = l^A[c_i],\\
    \beta_i & = l^B[c_i].
\end{align*}
This completes the proof of Equation \ref{eq:loss_inequality}, and thus optimising FES on a single instance $l$ is a convex problem. As the sum of convex functions is a convex function, optimising FES on a full batch $L$ is also a convex problem. Therefore, FES is a convex optimisation problem.

\subsection{Convolutional Feature Extractor Stacking}\label{sec:ConFES}

\begin{figure}[t]
    \centering
    \includegraphics[width=\textwidth]{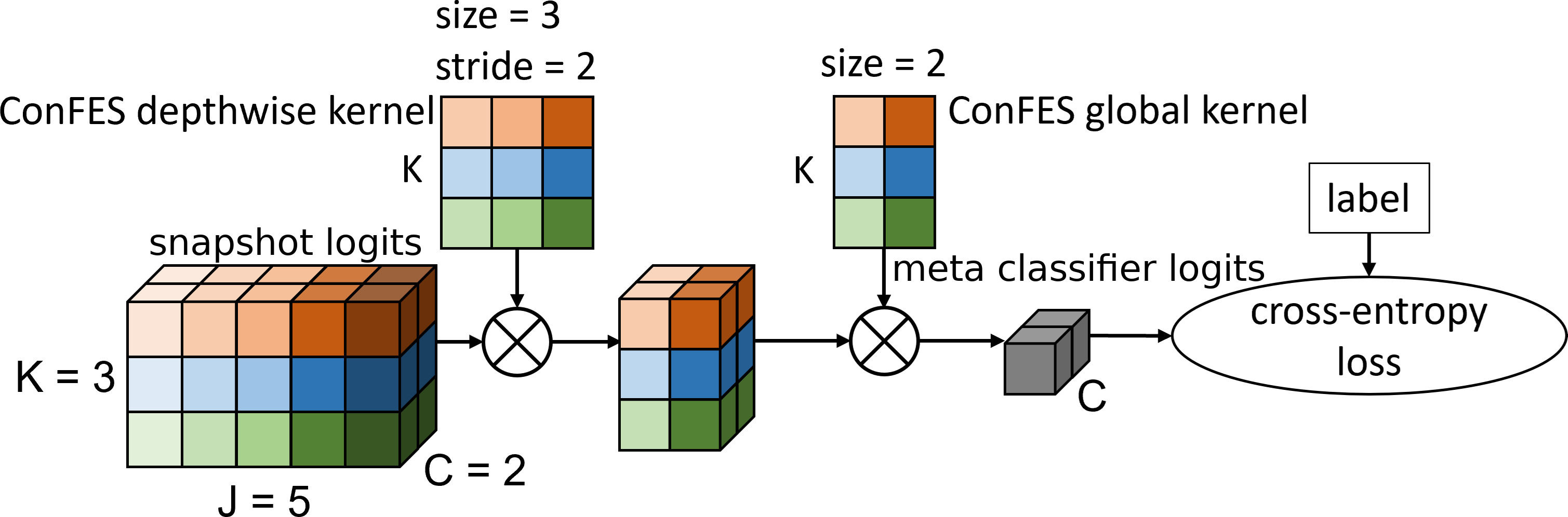}
    \caption{ConFES replaces the flat kernel of FES with a two-level kernel hierarchy. The base-level kernel is a one-dimensional depthwise, i.e., feature-extractor-wise, convolutional kernel, with predefined kernel and stride sizes. The high-level kernel is global like the one in FES but applied to the output of the base-level kernel, which requires substantially fewer parameters.}
    \label{fig:ConFES_metaclassifier}
\end{figure}

The basic FES approach does not exploit the temporal relation between logits obtained from adjacent snapshots produced during fine-tuning. Convolutional FES (ConFES) replaces the global kernel of FES with a kernel hierarchy, as shown in Figure \ref{fig:ConFES_metaclassifier}, to treat the collection of logits as a time series. The hierarchy comprises one or more lower-level one-dimensional depthwise convolutional kernels and a top-level global kernel. The depthwise kernels condense the logit output sequence from each extractor's snapshots into a 1D feature map, while keeping the extractors separate, and the global kernel summarises the feature maps produced by the lower-level kernels.

ConFES is motivated by the assumption that when each extractor is fine-tuned on the support set, it undergoes gradual changes between iterations, and the logits output by sequentially saved snapshots can be considered a time series. Therefore, 1D convolutions can be used to discern informative patterns in the time series data and compute feature maps, which are smaller in size than the raw logit time series, and therefore require fewer parameters in the global kernel than standard FES.

Given $K$ extractors and $J$ snapshots for each extractor, FES requires $K \times J$ parameters. Assuming a two-level ConFES hierarchy, with a base-level convolutional kernel of size $J_b$ and stride $T$, the feature map for each extractor will be of length $J_m = \frac{J - J_b}{T} + 1$, leading to a global kernel size of $K \times (\frac{J - J_b}{T} + 1)$. Including the $K \times J_b$ parameters in the convolutional kernel, ConFES contains $K \times (\frac{J - J_b}{T} + 1 + J_b)$ parameters. In practice, it can generally be assumed that $J \gg 1$: a two-level ConFES architecture should be configured so that $J \gg J_b \geq T \gg 1$ in order to cover all snapshots with significantly fewer parameters than FES.

ConFES utilises the sequential relation of each extractor's snapshots through its lower-level 1D depthwise convolutional layers and exhibits substantially fewer parameters than FES, making it less prone to overfitting. Note that Figure \ref{fig:ConFES_metaclassifier} is simplified for demonstration purposes and does not reflect well that ConFES maintains fewer parameters; for a practical example of ConFES kernels, please refer to Figure~\ref{fig:ConFES_heatmap}.

\subsection{Regularised Feature Extractor Stacking}\label{sec:ReFES}

\begin{figure}[t]
    \centering
    \includegraphics[width=\textwidth]{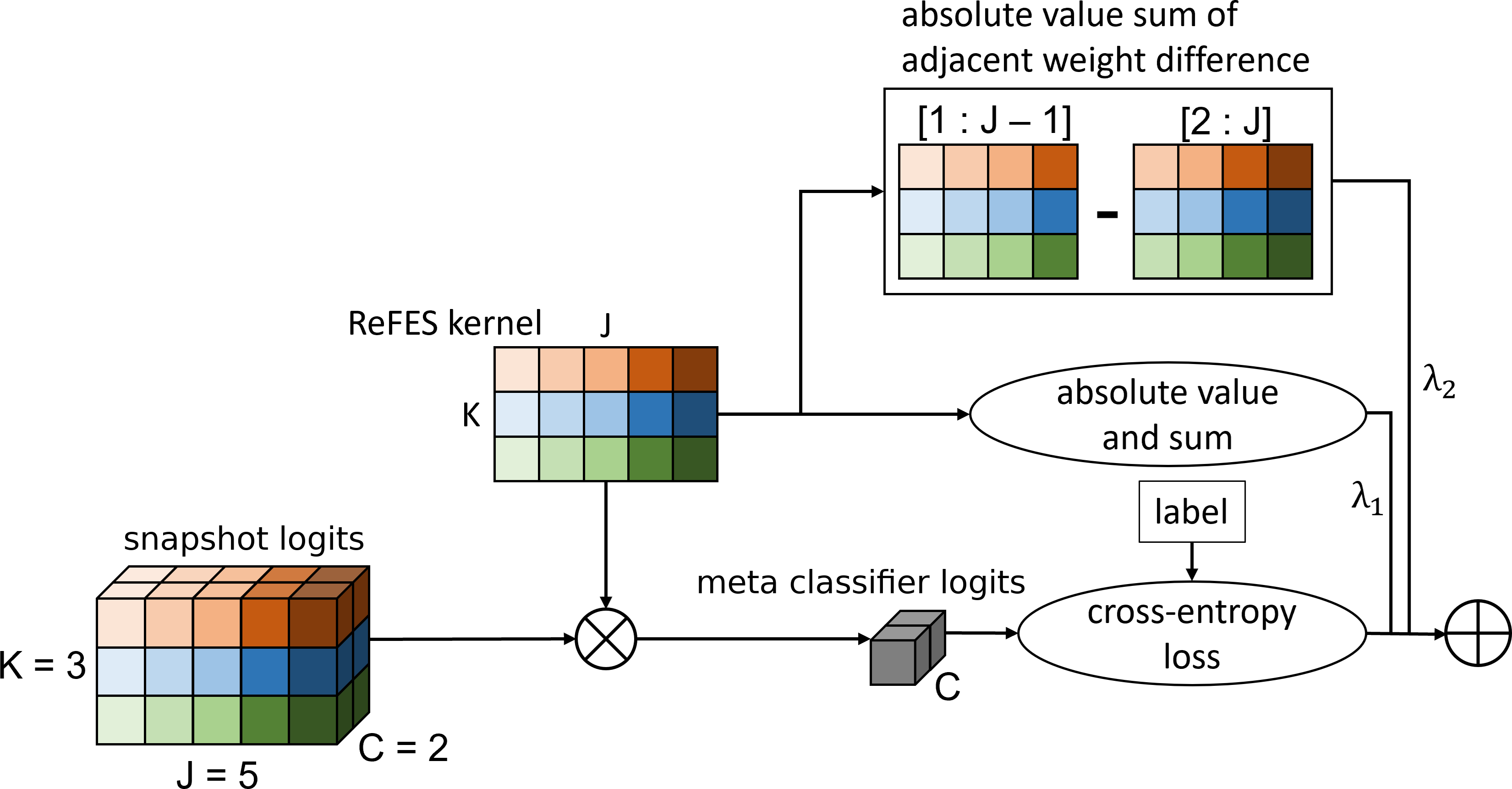}
    \caption{ReFES uses the same global kernel as FES and applies fused lasso regularisation to the kernel's training process. Fused lasso drives each individual weight towards zero with a regularisation strength of $\lambda_1$ and applies depthwise smoothing to the weight matrix by penalising the weight difference between adjacent snapshots with a regularisation strength of $\lambda_2$.}
    \label{fig:ReFES_metaclassifier}
\end{figure}

To combat overfitting, an alternative to reducing the number of parameters is to perform regularisation. Regularised FES (ReFES) introduces fused lasso regularisation~\citep{tibshirani2005sparsity} to the stacking classifier used in FES, as shown in Figure \ref{fig:ReFES_metaclassifier}. Non-zero weights are penalised with a strength of $\lambda_1$, and each feature-extractor-wise weight sequence is smoothed with a strength of $\lambda_2$. The loss is a combination of cross-entropy loss and depthwise fused lasso loss, as formulated in Equation \ref{eq:ReFES}, given $K$ extractors, $J$ snapshots per extractor, and a 2D global kernel $W$ of shape $K \times J$.
\begin{equation}\label{eq:ReFES}
    \ell = \ell_{\text{cross-entropy}} + \lambda_1 \sum_{k = 1}^K \sum_{j = 1}^J \|W_k^j\| + \lambda_2 \sum_{k = 1}^K \sum_{j = 1}^{J - 1} \|W_k^j - W_k^{j + 1}\|.
\end{equation}
In addition to encouraging sparse weights like standard lasso, fused lasso also encourages smaller differences between adjacent weights~\citep{tibshirani2005sparsity}. Each extractor's snapshots are ordered by their fine-tuning iterations, and adjacent snapshots are likely to be similar. By applying fused lasso regularisation, differences between adjacent weights are penalised, and weight sequences are smoothed.

The stratified two-fold splits $S_1$ and $S_2$ can be used to select appropriate $\lambda_1$ and $\lambda_2$ values for a few-shot episode. In the spirit of grid search with cross-validation, a ReFES stacking classifier is trained on the logits of one split, e.g., $L_K^J[S_1]$, and tested on the logits of the other split, e.g., $L_K^J[S_2]$. Different values for $\lambda_1$ and $\lambda_2$ can be explored and the best configuration selected based on the combined accuracy on the two folds. This configuration is then used to train a newly initialised ReFES stacking classifier on the full set of cross-validation logits $L_K^J[CV]$, and this stacking classifier is used to label the query set instances Q based on their logits $L_K^J[Q]$.

\subsection{Handling single-instance classes}\label{sec:single_instance}

Meta-Dataset's sampling scheme~\citep{Meta-Dataset2020Triantafillou} sometimes produces support sets containing single-instance classes. During cross-validation, single-instance classes need to be removed: if a class' only instance is in the test split $S^{test}$, then the training split $S^{train}$ will have no instance of that class. FES and its variants can train their stacking classifiers on a subset of the support classes $C_{sub} \leq C$, because their kernels only encode the weights of the snapshots, and are inherently independent of the number of classes $C$. In Figures \ref{fig:FES_metaclassifier}, \ref{fig:ConFES_metaclassifier}, and \ref{fig:ReFES_metaclassifier}, $C$ can simply be replaced by $C_{sub}$ during training.

Given a strict one-shot problem, where all classes exhibit exactly one instance, FES cross-validation is infeasible, as all classes need to be removed during cross-validation, leading to $L_K^J[CV] = \varnothing$. Therefore, support logits obtained from ordinary fine-tuning need to be used in place of cross-validation logits, i.e., $L_K^J[S]$ is used to train the FES classifier $W$ instead of using $L_K^J[CV]$.

\section{Experimental setup}\label{sec:experiment}

To evaluate FES and its variants on the Meta-Dataset benchmark described in Section \ref{sec:meta-dataset}, we use an extractor collection containing eight extractors, each independently pretrained on a Meta-Dataset source domain. In our primary set of experiments, all extractors are ResNet18 models~\citep{he2016deep} and identical to the source domain extractors used in the publication introducing URL~\citep{li2021universal}. Note that the extractors are trained on the training split of the source domain data only. The source domain validation split is used to select a trained checkpoint.

FES is compatible with any fine-tuning algorithm that is applicable to the individual extractors. In our experiments, we save a snapshot of each extractor before fine-tuning and save a snapshot after each iteration. We evaluate FES with three fine-tuning methods used by state-of-the-art CDFSL methods in the literature:

\begin{itemize}
    \item TSA~\citep{li2022cross}---matrix residual adaptors attached to convolutional layers, and a fully-connected layer to project feature vectors.
    \item URL~\citep{li2021universal}---only a fully-connected layer to project feature vectors.
    \item FLUTE~\citep{triantafillou2021learning}---scaling and shift factors of batch normalisation layers.
\end{itemize}

When performing each fine-tuning method for FES, we use the hyperparameters as stated in the source publications, including optimiser type, learning rate, number of iterations, etc., and we compare FES to each source method. The URL~\citep{li2021universal} and TSA~\citep{li2022cross} papers fine-tune their feature extractors for 40 iterations, leading to 41 FES snapshots per extractor. The FLUTE~\citep{triantafillou2021learning} paper fine-tunes its feature extractor for six iterations, leading to seven FES snapshots per extractor.

We adhere to the TSA, URL, and FLUTE papers when replicating and evaluating their methods as benchmarks. Pretrained universal extractors are obtained from the official repositories, and hyperparameter settings are consistent with the papers' specifications. Note that both the URL and TSA papers used the same URL-distilled universal extractor, and their difference is in fine-tuning, i.e., only fine-tuning a feature projection (URL) or additionally fine-tuning convolutional channel projections (TSA).

We use an LBFGS optimiser to train the stacking classifier, applying its default hyperparameters in the PyTorch library~\citep{NEURIPS2019_9015}, except that we utilise its line search function. A ridge regularisation of strength $1\mathrm{e}^{-2}$ is applied to FES and ConFES to make the LBFGS optimiser more numerically stable. Adjusting the regularisation strength up or down by an order of magnitude does not substantially affect classification accuracy.

\begin{table}[t]
\centering
\caption{Meta-Dataset episode statistics.}
\label{table:meta_dataset_statistics}
\begin{tabular}{lccccccccc}
\toprule
              & \multicolumn{3}{c}{support size} & \multicolumn{3}{c}{class count} & \multicolumn{3}{c}{mean shot} \\
Datasets      & min      & mean        & max     & min      & mean       & max     & min     & mean     & max      \\ \hline
ilsvrc\_2012  & 8        & 380.28      & 498     & 6        & 15.13      & 50      & 1       & 33.53    & 83.00    \\
omniglot      & 5        & 94.38       & 378     & 5        & 19.11      & 47      & 1       & 4.86     & 9.20     \\
aircraft      & 5        & 333.92      & 497     & 5        & 10.04      & 15      & 1       & 35.05    & 76.60    \\
cu\_birds     & 8        & 318.22      & 494     & 5        & 17.46      & 30      & 1       & 19.69    & 46.20    \\
dtd           & 5        & 290.94      & 498     & 5        & 6.00       & 7       & 1       & 48.64    & 97.20    \\
quickdraw     & 9        & 410.91      & 497     & 5        & 27.47      & 50      & 1       & 19.83    & 84.60    \\
fungi         & 6        & 350.79      & 494     & 5        & 26.38      & 50      & 1       & 16.31    & 65.43    \\
vgg\_flower   & 7        & 290.72      & 497     & 5        & 10.55      & 16      & 1       & 28.36    & 73.80    \\ \hline
traffic\_sign & 11       & 416.66      & 497     & 5        & 24.54      & 43      & 1       & 22.03    & 98.40    \\
mscoco        & 9        & 418.97      & 498     & 5        & 23.07      & 40      & 1       & 23.10    & 96.60    \\
mnist         & 5        & 325.46      & 498     & 5        & 7.52       & 10      & 1       & 44.33    & 99.60    \\
cifar10       & 7        & 318.78      & 498     & 5        & 7.47       & 10      & 1       & 44.16    & 99.40    \\
cifar100      & 9        & 409.28      & 497     & 5        & 27.32      & 50      & 1       & 19.74    & 84.60    \\
CropDisease   & 8        & 425.02      & 498     & 5        & 21.77      & 38      & 1       & 24.30    & 95.40    \\
EuroSAT       & 5        & 332.65      & 498     & 5        & 7.54       & 10      & 1       & 45.39    & 99.60    \\
ISIC          & 5        & 282.67      & 498     & 5        & 6.01       & 7       & 1       & 47.45    & 99.60    \\
ChestX        & 5        & 280.74      & 498     & 5        & 5.91       & 7       & 1       & 47.54    & 99.40    \\
Food101       & 7        & 420.13      & 498     & 5        & 26.99      & 50      & 1       & 20.53    & 98.40    \\
\bottomrule
\end{tabular}
\end{table}

Meta-Dataset's sampling randomness may cause one or two percent accuracy fluctuation of evaluated methods between different runs, as also stated in URL and TSA's code repositories~\citep{URL_TSA_FSL}. This fluctuation may exceed the 95\% confidence interval of most results, so to eliminate it, we sample 600 episodes from each domain once in Meta-Dataset. The sampled episodes are cached and then used to evaluate all CDFSL methods. In a dataset, the numbers of classes and instances are randomly sampled for each episode, which means that different episodes can contain different numbers of classes and instances. In an episode, the number of instances is randomly sampled for each class, which means that different classes can contain different numbers of instances, and episodes can be class-imbalanced. However, the query set is stratified and always contains 10 instances per class.

\citet{triantafillou2021learning} pointed out that Meta-Dataset instances need to be shuffled during sampling in case of datasets with particular ordering, e.g., traffic\_sign contains consecutive frames from the same video, but their shuffling solution was implemented as a moving window of size 1,000 for streams of instances of each class, which we found to be potentially insufficient, leading to approximately 1\% better accuracy in mscoco and 3\% better accuracy in ChestX than true random sampling. We found that a window size of 10,000 yielded virtually the same level of accuracy as true random sampling, but nevertheless use true random sampling in our experiments, i.e., instances in each class are fully randomised and have equal chance of being selected, and episodes are completely independent of each other. Statistics of our sampling run are shown in Table \ref{table:meta_dataset_statistics}. Using exactly the same sampled episodes for each learning scheme compared also allows us to perform a paired $t$-test on a per-dataset basis as a more sensitive statistical difference test than simply comparing two algorithms' mean accuracy and confidence intervals. In addition, we rank the algorithms and show their critical difference diagrams~\citep{demvsar2006statistical} in weak and strong generalisation.

Considering the complexity of the optimisation problem when learning the stacking classifier, it is worth noting that the FES and ReFES stacking classifiers each maintain $8 \times 41 = 328$ parameters if the extractors are fine-tuned for 40 iterations, and $8 \times 7 = 56$ parameters if the extractors are fine-tuned for 6 iterations.

ConFES is applied with a two-level hierarchy, i.e., a low-level depthwise 1D convolutional kernel and a high-level global kernel. For 40-iteration fine-tuning, the convolutional kernel has size $L = 9$ with stride $T = 4$, leading to a feature sequence/global kernel of length $9$. Consequently, ConFES has $8 \times 9 + 8 \times 9 = 144$ parameters in total. For 6-iteration fine-tuning, the convolutional kernel has size $3$ with stride $2$, leading to a global kernel of length $3$, and therefore ConFES contains $8 \times 3 + 8 \times 3 = 48$ parameters in total. All parameters are initialised with a constant $(1\mathrm{e}^{-3})^{\frac{1}{h}}$, where $h$ is the number of hierarchical levels in the stacking classifier. Therefore, FES and ReFES are initialised with $1\mathrm{e}^{-3}$, and a two-level ConFES hierarchy is initialised with $(1\mathrm{e}^{-3})^{\frac{1}{2}}$. This initialisation is deterministic and ensures that the product of weights from all levels is close to $1\mathrm{e}^{-3}$, which is small enough for optimisation to go in either direction, but also big enough to avoid exceedingly small derivatives in gradient-based optimisers.

To facilitate grid search for the $\lambda_1$ and $\lambda_2$ values of ReFES, a pool of eight potential values is provided for each hyperparameter: $1$, $1\mathrm{e}^{-1}$, $1\mathrm{e}^{-2}$, $1\mathrm{e}^{-3}$, $1\mathrm{e}^{-4}$, $1\mathrm{e}^{-5}$, $1\mathrm{e}^{-6}$, and $0$.

\section{Results}

We present CDFSL results of FES, ConFES, ReFES, and the competing methods URL, FLUTE, and a URL extractor with TSA fine-tuning, on the Meta-Dataset benchmark and show that FES and its variants advance the state of the art on this benchmark. We then visually analyse an example of trained FES, ConFES, and ReFES kernels. Lastly, we examine the ability of FES, ConFES, and ReFES to omit snapshots with their non-negative kernels.

\begin{table}[p]
\centering
\caption{Meta-Dataset results with TSA fine-tuning}
\label{table:results_tsa}
\begin{tabular}{lcc@{\hspace{0.1cm}}cc@{\hspace{0.1cm}}cc@{\hspace{0.1cm}}c}
\toprule
Dataset       & TSA                   & \multicolumn{2}{c}{FES}           & \multicolumn{2}{c}{ConFES}        & \multicolumn{2}{c}{ReFES} \\ \hline
ilsvrc\_2012  & \textBF{56.8$\pm$1.1} & 56.2$\pm$1.1          & $\bullet$ & 56.3$\pm$1.1          & $\bullet$ & 56.2$\pm$1.2           & $\bullet$ \\
omniglot      & \textBF{95.0$\pm$0.4} & 93.3$\pm$0.6          & $\bullet$ & 93.3$\pm$0.7          & $\bullet$ & 93.6$\pm$0.6           & $\bullet$ \\
aircraft      & \textBF{88.4$\pm$0.5} & 87.6$\pm$0.8          & $\bullet$ & 87.9$\pm$0.7          & $\bullet$ & 87.9$\pm$0.8           & $\bullet$ \\
cu\_birds     & \textBF{81.5$\pm$0.7} & 79.9$\pm$0.8          & $\bullet$ & 80.0$\pm$0.8          & $\bullet$ & 79.8$\pm$0.9           & $\bullet$ \\
dtd           & \textBF{77.1$\pm$0.7} & 76.2$\pm$0.8          & $\bullet$ & 76.3$\pm$0.8          & $\bullet$ & 76.4$\pm$0.8           & $\bullet$ \\
quickdraw     & 82.0$\pm$0.6          & 83.4$\pm$0.6          & $\circ$   & \textBF{83.5$\pm$0.6} & $\circ$   & 83.4$\pm$0.6           & $\circ$   \\
fungi         & 68.3$\pm$1.1          & 69.4$\pm$1.1          & $\circ$   & \textBF{69.7$\pm$1.1} & $\circ$   & 69.6$\pm$1.1           & $\circ$   \\
vgg\_flower   & \textBF{92.1$\pm$0.5} & 91.9$\pm$0.7          &           & 91.9$\pm$0.7          &           & \textBF{92.1$\pm$0.6}  &           \\ \hline
mean WG acc   & \textBF{80.15}        & \multicolumn{2}{c}{79.74}         & \multicolumn{2}{c}{79.86}         & \multicolumn{2}{c}{79.88}          \\
mean WG rank  & 2.58                  & \multicolumn{2}{c}{2.50}          & \multicolumn{2}{c}{\textBF{2.45}} & \multicolumn{2}{c}{2.47}           \\ \hline
traffic\_sign & 82.8$\pm$0.9          & 84.9$\pm$1.0          & $\circ$   & 85.1$\pm$1.0          & $\circ$   & \textBF{85.2$\pm$1.0}  & $\circ$   \\
mscoco        & 53.8$\pm$1.1          & 54.1$\pm$1.0          & $\circ$   & \textBF{54.4$\pm$1.0} & $\circ$   & \textBF{54.4$\pm$1.0}  & $\circ$   \\
mnist         & 96.6$\pm$0.4          & 97.1$\pm$0.5          & $\circ$   & 97.1$\pm$0.5          & $\circ$   & \textBF{97.2$\pm$0.5}  & $\circ$   \\
cifar10       & \textBF{79.9$\pm$0.8} & 78.1$\pm$0.9          & $\bullet$ & 78.2$\pm$0.9          & $\bullet$ & 78.8$\pm$0.9           & $\bullet$ \\
cifar100      & 70.3$\pm$1.0          & 70.4$\pm$1.1          &           & 70.6$\pm$1.0          & $\circ$   & \textBF{70.8$\pm$1.0}  & $\circ$   \\
CropDisease   & 84.4$\pm$0.8          & 88.1$\pm$0.7          & $\circ$   & \textBF{88.3$\pm$0.7} & $\circ$   & \textBF{88.3$\pm$0.7}  & $\circ$   \\
EuroSAT       & \textBF{89.6$\pm$0.5} & 88.8$\pm$0.6          & $\bullet$ & 89.1$\pm$0.6          & $\bullet$ & 89.3$\pm$0.6           &           \\
ISIC          & 48.4$\pm$0.9          & \textBF{49.5$\pm$0.9} & $\circ$   & 49.3$\pm$0.9          & $\circ$   & 48.9$\pm$0.9           & $\circ$   \\
ChestX        & 27.2$\pm$0.6          & 27.7$\pm$0.6          & $\circ$   & 27.7$\pm$0.6          & $\circ$   & \textBF{27.8$\pm$0.6}  & $\circ$   \\
Food101       & 53.4$\pm$1.2          & 55.2$\pm$1.1          & $\circ$   & \textBF{55.5$\pm$1.1} & $\circ$   & 55.2$\pm$1.1           & $\circ$   \\ \hline
mean SG acc   & 68.64                 & \multicolumn{2}{c}{69.39}         & \multicolumn{2}{c}{69.53}         & \multicolumn{2}{c}{\textBF{69.59}} \\
mean SG rank  & 2.85                  & \multicolumn{2}{c}{2.49}          & \multicolumn{2}{c}{2.34}          & \multicolumn{2}{c}{\textBF{2.32}} \\
\bottomrule
\end{tabular}
\end{table}

\begin{table}[p]
\centering
\caption{Statistically significant number of wins of column algorithm over row algorithm using paired $t$-test results with TSA fine-tuning}
\label{table:ttest_tsa}
\begin{tabular}{c@{\hspace{0.1cm}}c@{\hspace{0.1cm}}c@{\hspace{0.1cm}}c@{\hspace{0.1cm}}c}
\toprule
WG     & TSA & FES & ConFES & ReFES \\ \hline
TSA    & -   & 2   & 2      & 2     \\
FES    & 5   & -   & 4      & 3     \\
ConFES & 5   & 0   & -      & 2     \\
ReFES  & 5   & 1   & 1      & -     \\
\bottomrule
\end{tabular}
\begin{tabular}{c@{\hspace{0.1cm}}c@{\hspace{0.1cm}}c@{\hspace{0.1cm}}c@{\hspace{0.1cm}}c}
\toprule
SG     & TSA & FES & ConFES & ReFES \\ \hline
TSA    & -   & 7   & 8      & 8     \\
FES    & 2   & -   & 6      & 6     \\
ConFES & 2   & 0   & -      & 3     \\
ReFES  & 1   & 1   & 2      & -     \\
\bottomrule
\end{tabular}
\end{table}

\begin{figure}[p]
    \centering
    \includegraphics[width=\textwidth]{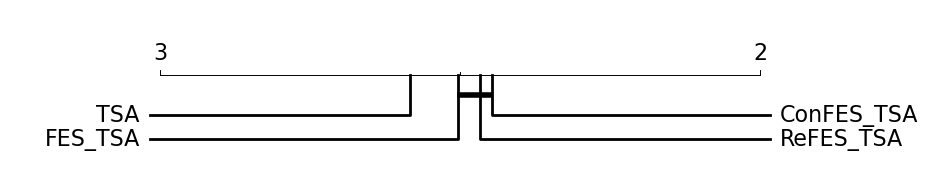}
    \caption{TSA weak generalisation critical difference diagram}
    \label{fig:tsa_weak_generalisation}
    \includegraphics[width=\textwidth]{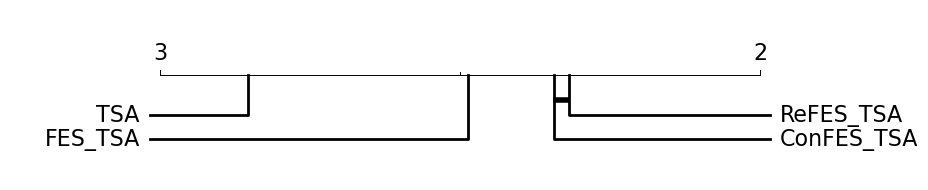}
    \caption{TSA strong generalisation critical difference diagram}
    \label{fig:tsa_strong_generalisation}
\end{figure}

\begin{table}[p]
\centering
\caption{Meta-Dataset results with URL fine-tuning}
\label{table:results_url}
\begin{tabular}{lcc@{\hspace{0.1cm}}cc@{\hspace{0.1cm}}cc@{\hspace{0.1cm}}c}
\toprule
Dataset       & URL                   & \multicolumn{2}{c}{FES}           & \multicolumn{2}{c}{ConFES}        & \multicolumn{2}{c}{ReFES}         \\ \hline
ilsvrc\_2012  & \textBF{56.6$\pm$1.1} & 56.1$\pm$1.1          & $\bullet$ & 56.1$\pm$1.1          & $\bullet$ & 55.9$\pm$1.2           & $\bullet$ \\
omniglot      & \textBF{94.5$\pm$0.4} & 93.1$\pm$0.6          & $\bullet$ & 93.4$\pm$0.6          & $\bullet$ & 93.5$\pm$0.6           & $\bullet$ \\
aircraft      & \textBF{87.7$\pm$0.5} & 87.1$\pm$0.8          & $\bullet$ & 87.5$\pm$0.7          &           & 87.5$\pm$0.7           &           \\
cu\_birds     & \textBF{80.7$\pm$0.7} & 79.1$\pm$0.8          & $\bullet$ & 79.2$\pm$0.8          & $\bullet$ & 79.0$\pm$0.8           & $\bullet$ \\
dtd           & \textBF{76.1$\pm$0.6} & 75.0$\pm$0.8          & $\bullet$ & 75.1$\pm$0.8          & $\bullet$ & 74.9$\pm$0.8           & $\bullet$ \\
quickdraw     & 82.0$\pm$0.6          & \textBF{83.2$\pm$0.6} & $\circ$   & \textBF{83.2$\pm$0.6} & $\circ$   & 83.1$\pm$0.6           & $\circ$   \\
fungi         & 69.5$\pm$1.1          & 69.6$\pm$1.1          &           & \textBF{69.8$\pm$1.1} &           & 69.6$\pm$1.1           &           \\
vgg\_flower   & \textBF{91.4$\pm$0.5} & 90.7$\pm$0.7          & $\bullet$ & 90.5$\pm$0.7          & $\bullet$ & 90.8$\pm$0.6           & $\bullet$ \\ \hline
mean WG acc   & \textBF{79.81}        & \multicolumn{2}{c}{79.24}         & \multicolumn{2}{c}{79.35}         & \multicolumn{2}{c}{79.29}          \\
mean WG rank  & 2.52                  & \multicolumn{2}{c}{2.50}          & \multicolumn{2}{c}{\textBF{2.47}} & \multicolumn{2}{c}{2.51}           \\ \hline
traffic\_sign & 62.6$\pm$1.2          & 65.2$\pm$1.2          & $\circ$   & \textBF{65.3$\pm$1.2} & $\circ$   & 65.0$\pm$1.2           & $\circ$   \\
mscoco        & 52.7$\pm$1.0          & 52.6$\pm$1.0          &           & 52.7$\pm$1.0          &           & \textBF{52.8$\pm$1.0}  &           \\
mnist         & 94.6$\pm$0.4          & 96.3$\pm$0.5          & $\circ$   & 96.4$\pm$0.5          & $\circ$   & \textBF{96.5$\pm$0.5}  & $\circ$   \\
cifar10       & 71.4$\pm$0.8          & 71.7$\pm$0.8          &           & \textBF{71.9$\pm$0.8} &           & \textBF{71.9$\pm$0.8}  &           \\
cifar100      & 62.6$\pm$1.1          & 62.7$\pm$1.1          &           & 62.7$\pm$1.1          &           & \textBF{62.8$\pm$1.1}  &           \\
CropDisease   & 80.5$\pm$0.8          & 87.2$\pm$0.7          & $\circ$   & 87.2$\pm$0.7          & $\circ$   & \textBF{87.4$\pm$0.7}  & $\circ$   \\
EuroSAT       & \textBF{86.6$\pm$0.5} & 86.1$\pm$0.6          & $\bullet$ & 86.1$\pm$0.6          & $\bullet$ & 86.3$\pm$0.6           &           \\
ISIC          & 45.5$\pm$0.8          & 48.4$\pm$0.9          & $\circ$   & 48.3$\pm$0.9          & $\circ$   & \textBF{48.9$\pm$0.9}  & $\circ$   \\
ChestX        & 26.5$\pm$0.6          & \textBF{27.2$\pm$0.6} & $\circ$   & 27.1$\pm$0.6          & $\circ$   & 27.1$\pm$0.6           & $\circ$   \\
Food101       & 51.9$\pm$1.1          & \textBF{53.9$\pm$1.1} & $\circ$   & \textBF{53.9$\pm$1.1} & $\circ$   & \textBF{53.9$\pm$1.1}  & $\circ$   \\ \hline
mean SG acc   & 63.49                 & \multicolumn{2}{c}{65.13}         & \multicolumn{2}{c}{65.16}         & \multicolumn{2}{c}{\textBF{65.26}} \\
mean SG rank  & 2.96                  & \multicolumn{2}{c}{2.40}          & \multicolumn{2}{c}{2.34}          & \multicolumn{2}{c}{\textBF{2.31}} \\
\bottomrule
\end{tabular}
\end{table}

\begin{table}[p]
\centering
\caption{Statistically significant number of wins of column algorithm over row algorithm using paired $t$-test results with URL fine-tuning}
\label{table:ttest_url}
\begin{tabular}{c@{\hspace{0.1cm}}c@{\hspace{0.1cm}}c@{\hspace{0.1cm}}c@{\hspace{0.1cm}}c}
\toprule
WG     & URL & FES & ConFES & ReFES \\ \hline
URL    & -   & 1   & 1      & 1     \\
FES    & 6   & -   & 3      & 2     \\
ConFES & 5   & 1   & -      & 1     \\
ReFES  & 5   & 3   & 3      & -     \\
\bottomrule
\end{tabular}
\begin{tabular}{c@{\hspace{0.1cm}}c@{\hspace{0.1cm}}c@{\hspace{0.1cm}}c@{\hspace{0.1cm}}c}
\toprule
SG     & URL & FES & ConFES & ReFES \\ \hline
URL    & -   & 6   & 6      & 6     \\
FES    & 1   & -   & 3      & 5     \\
ConFES & 1   & 0   & -      & 3     \\
ReFES  & 0   & 1   & 1      & -     \\
\bottomrule
\end{tabular}
\end{table}

\begin{figure}[p]
    \centering
    \includegraphics[width=\textwidth]{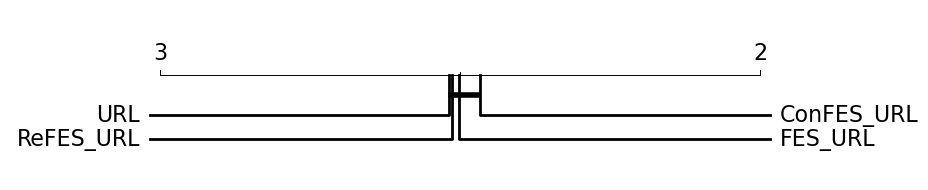}
    \caption{URL weak generalisation critical difference diagram ($p > \alpha$)}
    \label{fig:url_weak_generalisation}
    \includegraphics[width=\textwidth]{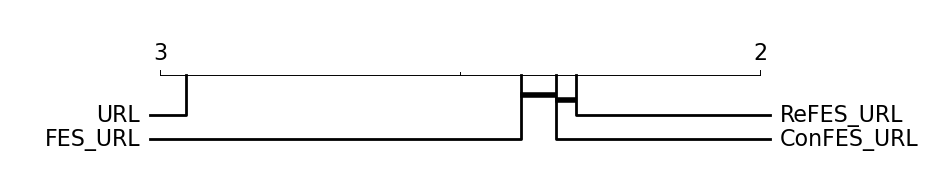}
    \caption{URL strong generalisation critical difference diagram}
    \label{fig:url_strong_generalisation}
\end{figure}

\begin{table}[p]
\centering
\caption{Meta-Dataset results with FLUTE fine-tuning}
\label{table:results_flute}
\begin{tabular}{lcc@{\hspace{0.1cm}}cc@{\hspace{0.1cm}}cc@{\hspace{0.1cm}}c}
\toprule
Dataset       & FLUTE                 & \multicolumn{2}{c}{FES}           & \multicolumn{2}{c}{ConFES}        & \multicolumn{2}{c}{ReFES}         \\ \hline
ilsvrc\_2012  & 50.2$\pm$1.1          & \textBF{54.1$\pm$1.1} & $\circ$   & \textBF{54.1$\pm$1.1}  & $\circ$   & 53.9$\pm$1.2          & $\circ$   \\
omniglot      & 93.9$\pm$0.5          & 94.3$\pm$0.6          & $\circ$   & 94.8$\pm$0.5           & $\circ$   & \textBF{94.9$\pm$0.5} & $\circ$   \\
aircraft      & 86.8$\pm$0.6          & \textBF{87.4$\pm$0.7} & $\circ$   & 87.1$\pm$0.9           &           & \textBF{87.4$\pm$0.7} & $\circ$   \\
cu\_birds     & \textBF{79.3$\pm$0.8} & 78.4$\pm$0.9          & $\bullet$ & 78.5$\pm$0.9           & $\bullet$ & 78.3$\pm$0.9          & $\bullet$ \\
dtd           & 68.8$\pm$0.8          & \textBF{74.3$\pm$0.9} & $\circ$   & \textBF{74.3$\pm$0.8}  & $\circ$   & 74.2$\pm$0.9          & $\circ$   \\
quickdraw     & 79.1$\pm$0.7          & 82.9$\pm$0.6          & $\circ$   & \textBF{83.0$\pm$0.6}  & $\circ$   & 82.7$\pm$0.6          & $\circ$   \\
fungi         & 59.4$\pm$1.2          & 68.7$\pm$1.1          & $\circ$   & \textBF{69.1$\pm$1.1}  & $\circ$   & 68.8$\pm$1.1          & $\circ$   \\
vgg\_flower   & 91.0$\pm$0.6          & 92.5$\pm$0.6          & $\circ$   & 92.5$\pm$0.6           & $\circ$   & \textBF{92.6$\pm$0.6} & $\circ$   \\ \hline
mean WG acc   & 76.06                 & \multicolumn{2}{c}{79.08}         & \multicolumn{2}{c}{\textBF{79.18}} & \multicolumn{2}{c}{79.10}         \\
mean WG rank  & 3.24                  & \multicolumn{2}{c}{2.27}          & \multicolumn{2}{c}{\textBF{2.22}}  & \multicolumn{2}{c}{2.27}          \\ \hline
traffic\_sign & 57.9$\pm$1.1          & 72.6$\pm$1.1          & $\circ$   & \textBF{72.8$\pm$1.1}  & $\circ$   & 72.3$\pm$1.1          & $\circ$   \\
mscoco        & 48.2$\pm$1.0          & 51.7$\pm$1.1          & $\circ$   & \textBF{51.8$\pm$1.1}  & $\circ$   & \textBF{51.8$\pm$1.1} & $\circ$   \\
mnist         & 95.7$\pm$0.4          & 97.4$\pm$0.4          & $\circ$   & \textBF{97.6$\pm$0.3}  & $\circ$   & 97.5$\pm$0.4          & $\circ$   \\
cifar10       & \textBF{78.6$\pm$0.7} & 75.2$\pm$0.9          & $\bullet$ & 75.2$\pm$0.9           & $\bullet$ & 75.1$\pm$0.9          & $\bullet$ \\
cifar100      & \textBF{67.5$\pm$1.0} & 67.2$\pm$1.1          & $\bullet$ & 67.2$\pm$1.0           &           & 67.0$\pm$1.1          & $\bullet$ \\
CropDisease   & 78.0$\pm$0.8          & \textBF{86.5$\pm$0.7} & $\circ$   & 86.4$\pm$0.7           & $\circ$   & \textBF{86.5$\pm$0.7} & $\circ$   \\
EuroSAT       & 81.6$\pm$0.6          & \textBF{88.3$\pm$0.6} & $\circ$   & \textBF{88.3$\pm$0.6}  & $\circ$   & 88.0$\pm$0.6          & $\circ$   \\
ISIC          & 46.1$\pm$1.0          & \textBF{48.7$\pm$1.0} & $\circ$   & \textBF{48.7$\pm$1.0}  & $\circ$   & 48.1$\pm$0.9          & $\circ$   \\
ChestX        & 26.3$\pm$0.5          & \textBF{27.8$\pm$0.6} & $\circ$   & 27.7$\pm$0.6           & $\circ$   & \textBF{27.8$\pm$0.6} & $\circ$   \\
Food101       & 45.7$\pm$1.1          & \textBF{52.1$\pm$1.1} & $\circ$   & 52.0$\pm$1.1           & $\circ$   & 52.0$\pm$1.1          & $\circ$   \\ \hline
mean SG acc   & 62.56                 & \multicolumn{2}{c}{66.75}         & \multicolumn{2}{c}{\textBF{66.77}} & \multicolumn{2}{c}{66.61}         \\
mean SG rank  & 3.24                  & \multicolumn{2}{c}{2.23}          & \multicolumn{2}{c}{\textBF{2.21}}  & \multicolumn{2}{c}{2.32}         \\
\bottomrule
\end{tabular}
\end{table}

\begin{table}[p]
\centering
\caption{Statistically significant number of wins of column algorithm over row algorithm using paired $t$-test results with FLUTE fine-tuning}
\label{table:ttest_flute}
\begin{tabular}{c@{\hspace{0.1cm}}c@{\hspace{0.1cm}}c@{\hspace{0.1cm}}c@{\hspace{0.1cm}}c}
\toprule
WG     & FLUTE & FES & ConFES & ReFES \\ \hline
FLUTE  & -     & 7   & 6      & 6     \\
FES    & 1     & -   & 7      & 5     \\
ConFES & 1     & 0   & -      & 0     \\
ReFES  & 1     & 0   & 7      & -     \\
\bottomrule
\end{tabular}
\begin{tabular}{c@{\hspace{0.1cm}}c@{\hspace{0.1cm}}c@{\hspace{0.1cm}}c@{\hspace{0.1cm}}c}
\toprule
SG     & FLUTE & FES & ConFES & ReFES \\ \hline
FLUTE  & -     & 7   & 8      & 8     \\
FES    & 3     & -   & 9      & 8     \\
ConFES & 2     & 0   & -      & 1     \\
ReFES  & 2     & 1   & 6      & -     \\
\bottomrule
\end{tabular}
\end{table}

\begin{figure}[p]
    \centering
    \includegraphics[width=\textwidth]{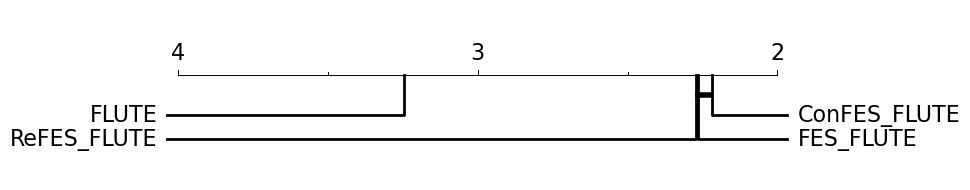}
    \caption{FLUTE weak generalisation critical difference diagram}
    \label{fig:flute_weak_generalisation}
    \includegraphics[width=\textwidth]{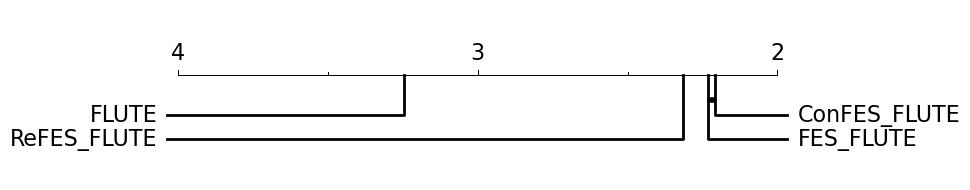}
    \caption{FLUTE strong generalisation critical difference diagram}
    \label{fig:flute_strong_generalisation}
\end{figure}

\subsection{Meta-Dataset results}

Results are organised by fine-tuning algorithms used, to provide a comparison between different CDFSL algorithms with the same fine-tuning scheme. The universal model of URL~\citep{li2021universal}, applied with TSA fine-tuning~\citep{li2022cross}, is the most recent and strongest CDFSL approach in the literature. Hence, we compare to this universal-model approach first, applying TSA fine-tuning in our FES methods as well in this comparison. Following that, we present experiments with the simpler fine-tuning approach used in the original URL~\citep{li2021universal} paper. Finally, we evaluate FLUTE~\citep{triantafillou2021learning} fine-tuning, which fine-tunes batch norm parameters only, and compare to the FLUTE universal template model.

Results with TSA fine-tuning are shown in Table \ref{table:results_tsa}, and paired $t$-test results based on the 600 individual accuracy values per dataset are shown in Table \ref{table:ttest_tsa}. Results with URL fine-tuning are shown in Tables \ref{table:results_url} and \ref{table:ttest_url}, and those with FLUTE fine-tuning are shown in Tables \ref{table:results_flute} and \ref{table:ttest_flute}.

In these tables, mean accuracy over 600 episodes and 95\% confidence intervals are shown for each algorithm and dataset, and weak and strong generalisation accuracy and ranks averaged over all individual episodes are listed below the datasets. The best result of each row is shown in \textBF{bold}. If a paired $t$-test between a FES algorithm and the corresponding universal model/template (in the leftmost column) returns a $p$ value less than $0.05$, the null hypothesis (that there is no statistically significant difference) is rejected, and the FES result is marked with either $\circ$ if it has higher accuracy, or $\bullet$ if its competitor has higher accuracy.

The tables showing paired $t$-test results are split by weak generalisation (the eight source domains) and strong generalisation (the ten target domains). Each value indicates the number of datasets where the algorithm in the value's column significantly outperforms the algorithm in its row according to the paired $t$-test.

Figures \ref{fig:tsa_weak_generalisation}, \ref{fig:tsa_strong_generalisation}, \ref{fig:url_weak_generalisation}, \ref{fig:url_strong_generalisation}, \ref{fig:flute_weak_generalisation}, and \ref{fig:flute_strong_generalisation} are critical difference diagrams produced by the Nemenyi test applied with $\alpha=0.05$, where algorithms are ranked using all relevant accuracy values ($8$ datasets $\times$ $600$ episodes for weak generalisation, and $10$ datasets $\times$ $600$ episodes for strong generalisation). A Friedman test is first performed on all algorithms with the same $\alpha$ to reject the null hypothesis. A Nemenyi test is then performed to group algorithms with no statistically significant difference into cliques via horizontal lines. Note that the Friedman $p$ value is greater than $\alpha$ for URL weak generalisation, i.e., Figure \ref{fig:url_weak_generalisation}, and the null hypothesis over all classifiers cannot be rejected in this case.

When using the same fine-tuning scheme, FES and its variants outperform their competitor CDFSL algorithms---building a universal model using knowledge distillation for URL and its TSA fine-tuning variant, and training a universal template with FiLM layers for FLUTE---in strong generalisation, where learning problems qualify as being cross-domain. The FES algorithms achieve better average accuracy and obtain more wins than losses in paired $t$-tests. They also rank higher than their competitors in the critical difference diagram.

Considering results with all three fine-tuning methods, the FES algorithms consistently outperform their competitors by a substantial margin on traffic\_sign, CropDisease, and Food101, while being outperformed on cifar10 and cifar100. This phenomenon may indicate that FES and its variants perform better in domains that are more specialised, while their competitors gain an edge on datasets more similar to ImageNet, such as the CIFAR datasets. This speculation is supported by the fact that the competitor methods artificially attach greater importance to ImageNet when their universal models are obtained~\citep{triantafillou2021learning,li2021universal,li2022cross}.

All three FES variants exhibit good CDFSL performance. Which variant is to be preferred depends on each specific use case: FES is the simplest and most versatile; ConFES maintains a smaller number of parameters and therefore a more manipulable search space; and ReFES uses regularisation to achieve smoother and more interpretable snapshot selections.

\begin{figure}[t]
    \centering
    \begin{subfigure}[b]{\textwidth}
        \includegraphics[width=\textwidth]{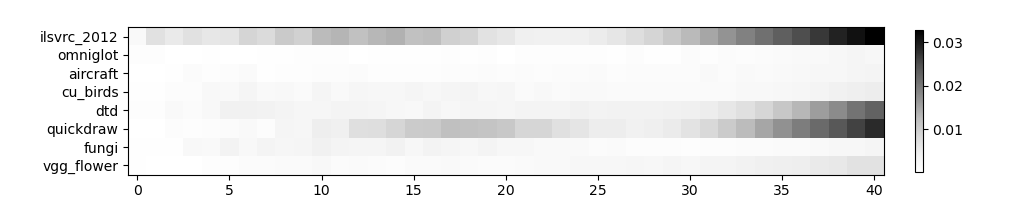}
    \end{subfigure}
    \caption{FES kernel for traffic\_sign}
    \label{fig:FES_heatmap}
    \begin{subfigure}[b]{0.4\textwidth}
        \centering
        \includegraphics[width=\textwidth]{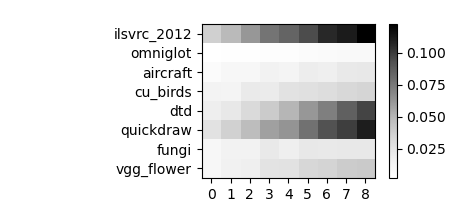}
        \caption{ConFES depthwise kernel}
        \label{fig:ConFES_heatmap_depthwise}
    \end{subfigure}
    \begin{subfigure}[b]{0.4\textwidth}
        \centering
        \includegraphics[width=\textwidth]{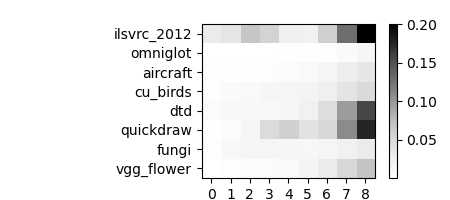}
        \caption{ConFES global kernel}
        \label{fig:ConFES_heatmap_global}
    \end{subfigure}
    \begin{subfigure}[b]{\textwidth}
        \includegraphics[width=\textwidth]{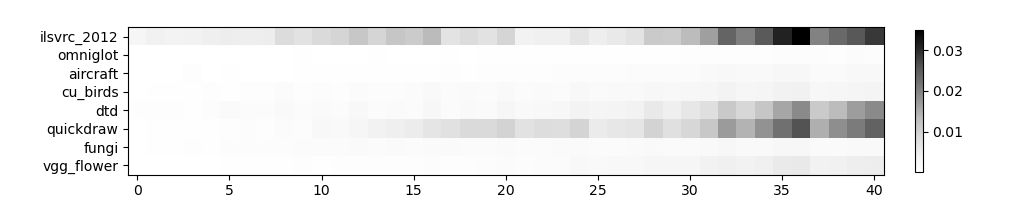}
        \caption{ConFES expanded kernel}
        \label{fig:ConFES_heatmap_expanded}
    \end{subfigure}
    \caption{ConFES kernels for traffic\_sign}
    \label{fig:ConFES_heatmap}
    \begin{subfigure}[b]{\textwidth}
        \includegraphics[width=\textwidth]{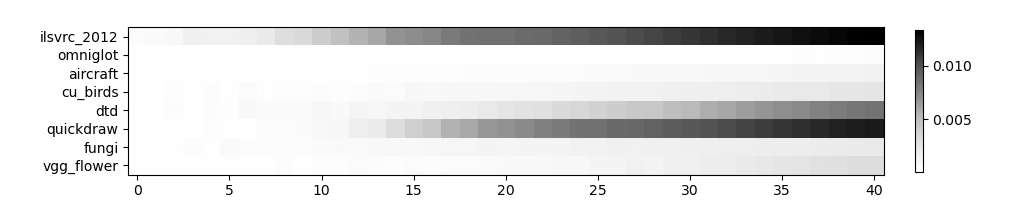}
    \end{subfigure}
    \caption{ReFES kernel for traffic\_sign}
    \label{fig:ReFES_heatmap}
\end{figure}

\subsection{Weight visualisation}

Weights of the FES, ConFES, and ReFES kernels after fine-tuning with TSA on traffic\_sign are visualised in Figures \ref{fig:FES_heatmap}, \ref{fig:ConFES_heatmap}, and \ref{fig:ReFES_heatmap}. The weights are averaged over 600 episodes.

ConFES maintains two kernels: a low-level depthwise 1D convolutional kernel (\ref{fig:ConFES_heatmap_depthwise}) and a high-level global kernel (\ref{fig:ConFES_heatmap_global}). The two kernels can be expanded back into a global kernel (\ref{fig:ConFES_heatmap_expanded}) for interpretation because the output of the convolutional kernel \ref{fig:ConFES_heatmap_depthwise} serves as direct input to the global kernel \ref{fig:ConFES_heatmap_global}, without any intermediate non-linear activation. Figure \ref{fig:ConFES_heatmap} demonstrates how ConFES emulates a 328-parameter FES kernel with only 144 parameters. The stepped pattern in the expanded ConFES kernel, where every fourth snapshot is assigned relatively greater weight than its neighbours, is an artefact of 1D convolution---with a kernel size of 9 and a stride size of 4, this pattern results from kernel overlaps.

FES determines that the fine-tuned ilsvrc\_2012 (ImageNet) and quickdraw extractors are the most prominent contributors to its predictions, indicated by the dark regions on the right end of these two extractors' rows in Figure \ref{fig:FES_heatmap}. ConFES and ReFES arrive at similar conclusions regarding contributors, but exhibit characteristics that reflect their classifiers' behaviours: ConFES shows stepped patterns due to 1D convolution as in Figure \ref{fig:ConFES_heatmap}; ReFES shows smoother weight changes due to fused lasso regularisation as in Figure \ref{fig:ReFES_heatmap}.

Additional heatmaps visualising kernel weights on the other target domains are in Appendix \ref{sec:appendix_heatmaps}, shown by Figures \ref{fig:appendix_start}-\ref{fig:appendix_end}.

\subsection{Snapshot omission}\label{sec:snapshot_omission}

\begin{table}[t]
\centering
\caption{Percentage of snapshots omitted by the stacking classifier}
\label{table:snapshot_omission_percentage}
\begin{tabular}{lc@{\hspace{0.25cm}}c@{\hspace{0.25cm}}c@{\hspace{0.25cm}}c@{\hspace{0.25cm}}c@{\hspace{0.25cm}}c@{\hspace{0.25cm}}c@{\hspace{0.25cm}}c@{\hspace{0.25cm}}c}
\toprule
              & \multicolumn{3}{c}{FES} & \multicolumn{3}{c}{ConFES} & \multicolumn{3}{c}{ReFES} \\
Dataset       & TSA    & URL   & FLUTE  & TSA     & URL    & FLUTE   & TSA    & URL    & FLUTE   \\ \hline
ilsvrc\_2012  & 72.2 & 72.4 & 65.9  & 58.1 & 56.0 & 33.5  & 38.2 & 38.3 & 40.0  \\
omniglot      & 51.2 & 51.0 & 45.5  & 42.7 & 43.5 & 36.2  & 22.1 & 27.0 & 24.0  \\
aircraft      & 71.0 & 72.5 & 58.6  & 50.6 & 48.3 & 32.5  & 37.0 & 37.1 & 30.7  \\
cu\_birds     & 73.2 & 71.3 & 68.3  & 62.5 & 58.4 & 48.8  & 46.6 & 45.2 & 41.8  \\
dtd           & 68.6 & 72.5 & 64.8  & 53.4 & 56.6 & 34.5  & 30.7 & 39.5 & 28.7  \\
quickdraw     & 69.0 & 70.0 & 69.1  & 58.6 & 57.7 & 34.5  & 31.7 & 36.5 & 42.3  \\
fungi         & 73.2 & 74.1 & 71.9  & 65.3 & 64.2 & 46.8  & 46.3 & 46.5 & 42.5  \\
vgg\_flower   & 64.2 & 66.3 & 46.7  & 40.2 & 42.5 & 26.3  & 24.4 & 28.6 & 24.3  \\ \hline
traffic\_sign & 75.6 & 79.1 & 81.6  & 61.9 & 73.3 & 74.5  & 29.7 & 55.6 & 70.5  \\
mscoco        & 79.2 & 76.8 & 70.2  & 60.6 & 55.3 & 38.1  & 35.0 & 34.8 & 39.4  \\
mnist         & 62.5 & 68.6 & 51.4  & 39.3 & 44.9 & 34.8  & 19.4 & 21.0 & 27.2  \\
cifar10       & 78.6 & 80.3 & 68.0  & 67.4 & 65.1 & 37.3  & 37.4 & 35.3 & 39.9  \\
cifar100      & 78.1 & 74.3 & 65.0  & 61.0 & 56.8 & 28.8  & 35.4 & 32.9 & 40.3  \\
CropDisease   & 73.4 & 73.7 & 52.2  & 51.2 & 48.8 & 20.6  & 23.9 & 23.1 & 26.2  \\
EuroSAT       & 75.6 & 74.8 & 61.6  & 62.4 & 63.3 & 32.1  & 28.2 & 27.3 & 29.6  \\
ISIC          & 70.1 & 72.1 & 67.0  & 58.6 & 62.1 & 40.0  & 25.7 & 33.7 & 25.6  \\
ChestX        & 85.2 & 82.9 & 77.3  & 75.4 & 67.8 & 44.8  & 44.3 & 44.2 & 42.2  \\
Food101       & 80.6 & 76.5 & 61.4  & 62.9 & 53.7 & 26.9  & 53.6 & 37.4 & 36.9  \\
\bottomrule
\end{tabular}
\end{table}

As FES kernels are constrained to be non-negative by clipping their weights with ReLU, some snapshots may have their corresponding weights set to $0$ after clipping, which means logits from these snapshots do not contribute to the aggregated meta logits, and these snapshots can be omitted, i.e., they do not need to be saved and are not used for inference.

Table \ref{table:snapshot_omission_percentage} shows the average percentage of snapshots omitted by a FES, ConFES, or ReFES stacking classifier, using TSA, URL, or FLUTE fine-tuning. Note that ConFES omission rates are computed using the expanded kernel, because zero values need to exist in the expanded kernel, instead of merely in one of ConFES' hierarchical kernels, for the corresponding snapshots to be omitted. A higher omission percentage is considered better because omitting snapshots saves storage space and inference computation. Among the three methods, FES achieves the highest percentage of omission, generally between 60\% and 80\%, followed by ConFES, which achieves 30\% to 70\% omission in general, while ReFES achieves the least amount of omission, mostly below 40\%. FES achieves higher omission rates than ConFES and ReFES, but trades off mean strong generalisation accuracy as shown in Tables \ref{table:results_tsa}, \ref{table:results_url}, and \ref{table:results_flute}.

\section{Ablation study}

\begin{table}[t]
\centering
\caption{Ablation results with TSA fine-tuning}
\label{table:ablation_tsa}
\begin{tabular}{lcccccc}
\toprule
              & \multicolumn{3}{c}{FES without cross-validation}                                 & \multicolumn{3}{c}{FES}                                    \\
Dataset       & first                & last        & all               & first                & last        & all               \\ \hline
ilsvrc\_2012  & 52.3$\pm$1.2          & 52.7$\pm$1.2 & 53.5$\pm$1.2 & 54.5$\pm$1.1          & \textBF{56.4$\pm$1.2} & 56.2$\pm$1.1          \\
omniglot      & \textBF{94.5$\pm$0.5} & 90.2$\pm$0.7 & 90.6$\pm$0.7 & \textBF{94.5$\pm$0.5} & 93.3$\pm$0.7          & 93.3$\pm$0.6          \\
aircraft      & 85.7$\pm$0.7          & 78.5$\pm$1.1 & 78.1$\pm$1.1 & 87.1$\pm$0.6          & \textBF{87.7$\pm$0.8} & 87.6$\pm$0.8          \\
cu\_birds     & 75.9$\pm$0.9          & 73.1$\pm$1.1 & 74.2$\pm$1.1 & 78.9$\pm$0.8          & 79.8$\pm$0.9          & \textBF{79.9$\pm$0.8} \\
dtd           & 72.2$\pm$0.8          & 75.5$\pm$0.9 & 75.9$\pm$0.8 & 72.9$\pm$0.8          & \textBF{76.8$\pm$0.8} & 76.2$\pm$0.8          \\
quickdraw     & 82.8$\pm$0.6          & 81.9$\pm$0.8 & 82.5$\pm$0.7 & 83.0$\pm$0.6          & \textBF{83.4$\pm$0.6} & \textBF{83.4$\pm$0.6} \\
fungi         & 64.6$\pm$1.2          & 57.8$\pm$1.3 & 59.1$\pm$1.3 & 69.2$\pm$1.1          & 69.2$\pm$1.1          & \textBF{69.4$\pm$1.1} \\
vgg\_flower   & 90.0$\pm$0.6          & 89.9$\pm$0.8 & 90.1$\pm$0.7 & 90.3$\pm$0.6          & \textBF{92.1$\pm$0.7} & 91.9$\pm$0.7          \\ \hline
mean WG acc   & 77.25                & 74.95       & 75.50       & 78.80                & \textBF{79.84}       & 79.74                \\ \hline
traffic\_sign & 48.9$\pm$1.1          & 84.8$\pm$1.0 & 85.0$\pm$1.0 & 49.2$\pm$1.1          & \textBF{85.8$\pm$0.9} & 84.9$\pm$1.0          \\
mscoco        & 45.9$\pm$1.1          & 49.8$\pm$1.0 & 51.9$\pm$1.0 & 48.3$\pm$1.0          & 53.6$\pm$1.0          & \textBF{54.1$\pm$1.0} \\
mnist         & 95.8$\pm$0.4          & 96.5$\pm$0.5 & 96.7$\pm$0.5 & 95.6$\pm$0.5          & \textBF{97.2$\pm$0.5} & 97.1$\pm$0.5          \\
cifar10       & 66.9$\pm$0.9          & 72.2$\pm$1.0 & 74.8$\pm$1.0 & 68.6$\pm$0.8          & \textBF{78.6$\pm$0.9} & 78.1$\pm$0.9          \\
cifar100      & 57.0$\pm$1.1          & 65.2$\pm$1.1 & 68.0$\pm$1.1 & 59.1$\pm$1.1          & \textBF{70.7$\pm$1.0} & 70.4$\pm$1.1          \\
CropDisease   & 81.9$\pm$0.8          & 87.0$\pm$0.7 & 87.8$\pm$0.7 & 82.6$\pm$0.7          & 88.0$\pm$0.7          & \textBF{88.1$\pm$0.7} \\
EuroSAT       & 81.3$\pm$0.6          & 87.7$\pm$0.7 & 88.5$\pm$0.6 & 81.5$\pm$0.6          & \textBF{89.5$\pm$0.6} & 88.8$\pm$0.6          \\
ISIC          & 46.2$\pm$0.8          & 47.0$\pm$0.9 & 47.8$\pm$0.9 & 46.4$\pm$0.8          & 47.9$\pm$0.9          & \textBF{49.5$\pm$0.9} \\
ChestX        & 25.0$\pm$0.5          & 27.2$\pm$0.6 & 27.6$\pm$0.6 & 24.7$\pm$0.5          & 27.2$\pm$0.6          & \textBF{27.7$\pm$0.6} \\
Food101       & 49.4$\pm$1.2          & 50.9$\pm$1.2 & 51.9$\pm$1.2 & 52.3$\pm$1.1          & 54.5$\pm$1.1          & \textBF{55.2$\pm$1.1} \\ \hline
mean SG acc   & 59.83                & 66.83       & 68.00       & 60.83                & 69.30                & \textBF{69.39}       \\
\bottomrule
\end{tabular}
\end{table}

\begin{table}[t]
\centering
\caption{Ablation results with URL fine-tuning}
\label{table:ablation_url}
\begin{tabular}{lcccccc}
\toprule
              & \multicolumn{3}{c}{FES without cross-validation}                                 & \multicolumn{3}{c}{FES}                                    \\
Dataset       & first                & last        & all               & first                & last        & all               \\ \hline
ilsvrc\_2012  & 52.3$\pm$1.2          & 52.6$\pm$1.2 & 52.8$\pm$1.2          & 54.5$\pm$1.1          & \textBF{56.3$\pm$1.2} & 56.1$\pm$1.1          \\
omniglot      & \textBF{94.5$\pm$0.5} & 90.1$\pm$0.7 & 90.8$\pm$0.7          & \textBF{94.5$\pm$0.5} & 93.0$\pm$0.7          & 93.1$\pm$0.6          \\
aircraft      & 85.7$\pm$0.7          & 82.0$\pm$1.1 & 82.3$\pm$1.0          & 87.1$\pm$0.6          & \textBF{87.3$\pm$0.8} & 87.1$\pm$0.8          \\
cu\_birds     & 75.9$\pm$0.9          & 73.8$\pm$1.0 & 74.4$\pm$1.0          & 78.9$\pm$0.8          & 79.0$\pm$0.9          & \textBF{79.1$\pm$0.8} \\
dtd           & 72.2$\pm$0.8          & 73.9$\pm$0.8 & 73.7$\pm$0.8          & 72.9$\pm$0.8          & \textBF{75.4$\pm$0.8} & 75.0$\pm$0.8          \\
quickdraw     & 82.8$\pm$0.6          & 82.1$\pm$0.7 & 82.3$\pm$0.7          & 83.0$\pm$0.6          & 83.1$\pm$0.6          & \textBF{83.2$\pm$0.6} \\
fungi         & 64.6$\pm$1.2          & 62.2$\pm$1.3 & 63.0$\pm$1.3          & 69.2$\pm$1.1          & 69.5$\pm$1.1          & \textBF{69.6$\pm$1.1} \\
vgg\_flower   & 90.0$\pm$0.6          & 89.7$\pm$0.8 & 89.5$\pm$0.7          & 90.3$\pm$0.6          & \textBF{90.9$\pm$0.7} & 90.7$\pm$0.7          \\ \hline
mean WG acc   & 77.25                & 75.80       & 76.10                & 78.80                & \textBF{79.31}       & 79.24                \\ \hline
traffic\_sign & 48.9$\pm$1.1          & 65.0$\pm$1.2 & 65.2$\pm$1.2          & 49.2$\pm$1.1          & \textBF{65.4$\pm$1.2} & 65.2$\pm$1.2          \\
mscoco        & 45.9$\pm$1.1          & 49.3$\pm$1.1 & 49.9$\pm$1.1          & 48.3$\pm$1.0          & 51.2$\pm$1.0          & \textBF{52.6$\pm$1.0} \\
mnist         & 95.8$\pm$0.4          & 96.3$\pm$0.5 & 96.5$\pm$0.5          & 95.6$\pm$0.5          & \textBF{96.4$\pm$0.5} & 96.3$\pm$0.5          \\
cifar10       & 66.9$\pm$0.9          & 69.8$\pm$0.9 & 70.6$\pm$0.9          & 68.6$\pm$0.8          & 71.1$\pm$0.8          & \textBF{71.7$\pm$0.8} \\
cifar100      & 57.0$\pm$1.1          & 60.8$\pm$1.2 & 61.6$\pm$1.2          & 59.1$\pm$1.1          & \textBF{62.8$\pm$1.1} & 62.7$\pm$1.1          \\
CropDisease   & 81.9$\pm$0.8          & 86.6$\pm$0.7 & 86.7$\pm$0.7          & 82.6$\pm$0.7          & 87.0$\pm$0.7          & \textBF{87.2$\pm$0.7} \\
EuroSAT       & 81.3$\pm$0.6          & 86.1$\pm$0.6 & \textBF{86.3$\pm$0.6} & 81.5$\pm$0.6          & 86.0$\pm$0.6          & 86.1$\pm$0.6          \\
ISIC          & 46.2$\pm$0.8          & 46.7$\pm$0.9 & 45.7$\pm$0.9          & 46.4$\pm$0.8          & 47.6$\pm$0.9          & \textBF{48.4$\pm$0.9} \\
ChestX        & 25.0$\pm$0.5          & 27.3$\pm$0.6 & \textBF{27.4$\pm$0.6} & 24.7$\pm$0.5          & 26.4$\pm$0.6          & 27.2$\pm$0.6          \\
Food101       & 49.4$\pm$1.2          & 49.5$\pm$1.2 & 50.2$\pm$1.2          & 52.3$\pm$1.1          & 53.1$\pm$1.1          & \textBF{53.9$\pm$1.1} \\ \hline
mean SG acc   & 59.83                & 63.74       & 64.01                & 60.83                & 64.70                & \textBF{65.13}       \\
\bottomrule
\end{tabular}
\end{table}

\begin{table}[t]
\centering
\caption{Ablation results with FLUTE fine-tuning}
\label{table:ablation_flute}
\begin{tabular}{lcccccc}
\toprule
              & \multicolumn{3}{c}{FES without cross-validation}                                 & \multicolumn{3}{c}{FES}                                    \\
Dataset       & first                & last        & all               & first                & last        & all               \\ \hline
ilsvrc\_2012  & 49.9$\pm$1.2 & 50.5$\pm$1.2 & 50.6$\pm$1.2 & 53.2$\pm$1.1          & 53.5$\pm$1.2          & \textBF{54.1$\pm$1.1} \\
omniglot      & 93.0$\pm$0.6 & 93.2$\pm$0.6 & 93.3$\pm$0.6 & 94.1$\pm$0.5          & 94.2$\pm$0.6          & \textBF{94.3$\pm$0.6} \\
aircraft      & 83.7$\pm$0.9 & 83.1$\pm$1.0 & 83.2$\pm$1.0 & 86.9$\pm$0.7          & 87.3$\pm$0.8          & \textBF{87.4$\pm$0.7} \\
cu\_birds     & 72.7$\pm$1.0 & 73.4$\pm$1.1 & 73.7$\pm$1.1 & 76.8$\pm$0.9          & 78.3$\pm$0.9          & \textBF{78.4$\pm$0.9} \\
dtd           & 72.0$\pm$0.8 & 73.9$\pm$0.9 & 73.9$\pm$0.9 & 72.5$\pm$0.8          & \textBF{74.5$\pm$0.9} & 74.3$\pm$0.9          \\
quickdraw     & 81.2$\pm$0.7 & 81.2$\pm$0.7 & 81.4$\pm$0.7 & 82.7$\pm$0.6          & 81.8$\pm$0.6          & \textBF{82.9$\pm$0.6} \\
fungi         & 60.4$\pm$1.3 & 59.0$\pm$1.3 & 59.3$\pm$1.3 & \textBF{68.7$\pm$1.1} & 67.5$\pm$1.1          & \textBF{68.7$\pm$1.1} \\
vgg\_flower   & 90.9$\pm$0.7 & 91.7$\pm$0.7 & 91.7$\pm$0.7 & 91.8$\pm$0.6          & \textBF{92.6$\pm$0.6} & 92.5$\pm$0.6          \\ \hline
mean WG acc   & 75.48       & 75.75       & 75.89       & 78.34                & 78.71                & \textBF{79.08}       \\ \hline
traffic\_sign & 53.0$\pm$1.1 & 71.7$\pm$1.1 & 71.0$\pm$1.1 & 53.2$\pm$1.1          & \textBF{72.9$\pm$1.1} & 72.6$\pm$1.1          \\
mscoco        & 44.5$\pm$1.1 & 49.0$\pm$1.1 & 49.0$\pm$1.1 & 47.1$\pm$1.0          & 51.6$\pm$1.1          & \textBF{51.7$\pm$1.1} \\
mnist         & 96.0$\pm$0.4 & 97.1$\pm$0.5 & 97.1$\pm$0.5 & 96.2$\pm$0.4          & \textBF{97.5$\pm$0.4} & 97.4$\pm$0.4          \\
cifar10       & 68.3$\pm$0.9 & 73.6$\pm$0.9 & 73.7$\pm$0.9 & 70.1$\pm$0.8          & \textBF{75.4$\pm$0.9} & 75.2$\pm$0.9          \\
cifar100      & 59.0$\pm$1.2 & 64.2$\pm$1.2 & 64.2$\pm$1.2 & 61.4$\pm$1.1          & \textBF{67.3$\pm$1.0} & 67.2$\pm$1.1          \\
CropDisease   & 83.1$\pm$0.8 & 86.1$\pm$0.7 & 85.6$\pm$0.7 & 84.2$\pm$0.7          & \textBF{86.5$\pm$0.7} & \textBF{86.5$\pm$0.7} \\
EuroSAT       & 86.2$\pm$0.6 & 88.2$\pm$0.6 & 88.1$\pm$0.6 & 86.1$\pm$0.6          & \textBF{88.5$\pm$0.6} & 88.3$\pm$0.6          \\
ISIC          & 48.2$\pm$0.9 & 45.0$\pm$0.9 & 45.0$\pm$0.9 & \textBF{48.8$\pm$0.9} & \textBF{48.8$\pm$0.9} & 48.7$\pm$1.0          \\
ChestX        & 26.2$\pm$0.5 & 27.5$\pm$0.6 & 27.4$\pm$0.6 & 25.9$\pm$0.6          & \textBF{27.8$\pm$0.6} & \textBF{27.8$\pm$0.6} \\
Food101       & 45.7$\pm$1.2 & 48.1$\pm$1.2 & 48.2$\pm$1.2 & 49.0$\pm$1.1          & 51.9$\pm$1.1          & \textBF{52.1$\pm$1.1} \\ \hline
mean SG acc   & 61.02       & 65.05       & 64.93       & 62.20                & \textBF{66.82}       & 66.75                \\
\bottomrule
\end{tabular}
\end{table}

We perform an ablation study by removing cross-validation from the framework and/or using only the first or last snapshots in fine-tuning. When cross-validation is not used, training logits for the ``stacking" classifier are extracted from the support set using snapshots fine-tuned on the entire support set, akin to how one-shot episodes are handled in Section \ref{sec:single_instance}. When using only the first or last snapshots, the stacking classifier is a degenerate weight kernel with a singleton dimension for fine-tuning iterations, simply containing one weight value for each extractor. Results are shown in Tables \ref{table:ablation_tsa}, \ref{table:ablation_url}, and \ref{table:ablation_flute}, organised by the fine-tuning algorithm used.

The results show that methods using cross-validation outperform their counterparts without cross-validation. Moreover, using all snapshots achieves better performance than using only the first or last snapshots in terms of mean strong generalisation performance for URL and TSA fine-tuning. For strong generalisation with FLUTE fine-tuning, using only the last snapshots leads to better performance. This could be due to the smaller number of fine-tuning iterations performed by FLUTE, as the last snapshots constitute a more substantial part of all snapshots. It is worth noting that cross-validation is helpful even when only using the first snapshots before any fine-tuning because the training logits are computed using a nearest centroid classifier, and cross-validation keeps the support instances for logit extraction separate from those used to compute the centroids, hence avoiding instance re-use and reducing overfitting.

\section{Heterogeneous extractors}

\begin{table}[t]
\centering
\caption{Results of replacing the ResNet18 ImageNet extractor with a Small EfficientNetV2 pretrained on the 21K version of ImageNet, while the other seven extractors remain the same.}
\label{table:efficientnetv2}
\begin{tabular}{lcccccc}
\toprule
              & \multicolumn{3}{c}{ResNet18}                              & \multicolumn{3}{c}{EfficientNetV2Small21K}                \\
Dataset       & FES                  & ConFES               & ReFES       & FES                  & ConFES               & ReFES       \\ \hline
ilsvrc\_2012  & 56.1$\pm$1.1          & 56.1$\pm$1.1 & 55.9$\pm$1.2          & 63.5$\pm$1.0          & \textBF{63.7$\pm$1.0} & 63.1$\pm$1.1          \\
omniglot      & 93.1$\pm$0.6          & 93.4$\pm$0.6 & \textBF{93.5$\pm$0.6} & 93.0$\pm$0.7          & \textBF{93.5$\pm$0.6} & 93.3$\pm$0.6          \\
aircraft      & 87.1$\pm$0.8          & 87.5$\pm$0.7 & 87.5$\pm$0.7          & 87.4$\pm$0.8          & \textBF{87.7$\pm$0.7} & 87.5$\pm$0.8          \\
cu\_birds     & 79.1$\pm$0.8          & 79.2$\pm$0.8 & 79.0$\pm$0.8          & 80.6$\pm$0.8          & \textBF{80.9$\pm$0.8} & 80.4$\pm$0.9          \\
dtd           & 75.0$\pm$0.8          & 75.1$\pm$0.8 & 74.9$\pm$0.8          & 81.5$\pm$0.8          & \textBF{81.7$\pm$0.7} & 81.0$\pm$0.8          \\
quickdraw     & 83.2$\pm$0.6          & 83.2$\pm$0.6 & 83.1$\pm$0.6          & \textBF{83.3$\pm$0.6} & \textBF{83.3$\pm$0.6} & 83.2$\pm$0.6          \\
fungi         & 69.6$\pm$1.1          & 69.8$\pm$1.1 & 69.6$\pm$1.1          & 70.0$\pm$1.1          & \textBF{70.2$\pm$1.1} & 70.0$\pm$1.1          \\
vgg\_flower   & 90.7$\pm$0.7          & 90.5$\pm$0.7 & 90.8$\pm$0.6          & 96.2$\pm$0.5          & \textBF{96.9$\pm$0.3} & 96.2$\pm$0.5          \\ \hline
mean WG acc   & 79.24                & 79.35       & 79.29                & 81.94                & \textBF{82.24}       & 81.84                \\ \hline
traffic\_sign & 65.2$\pm$1.2          & 65.3$\pm$1.2 & 65.0$\pm$1.2          & \textBF{65.6$\pm$1.1} & 65.5$\pm$1.1          & 65.4$\pm$1.2          \\
mscoco        & 52.6$\pm$1.0          & 52.7$\pm$1.0 & 52.8$\pm$1.0          & 60.7$\pm$1.0          & \textBF{60.8$\pm$1.0} & 60.0$\pm$1.0          \\
mnist         & 96.3$\pm$0.5          & 96.4$\pm$0.5 & 96.5$\pm$0.5          & 96.3$\pm$0.5          & 96.5$\pm$0.5          & \textBF{96.6$\pm$0.5} \\
cifar10       & 71.7$\pm$0.8          & 71.9$\pm$0.8 & 71.9$\pm$0.8          & 82.9$\pm$0.7          & \textBF{83.0$\pm$0.8} & 82.7$\pm$0.8          \\
cifar100      & 62.7$\pm$1.1          & 62.7$\pm$1.1 & 62.8$\pm$1.1          & 73.9$\pm$1.0          & \textBF{74.0$\pm$1.0} & 73.8$\pm$1.0          \\
CropDisease   & 87.2$\pm$0.7          & 87.2$\pm$0.7 & 87.4$\pm$0.7          & \textBF{89.6$\pm$0.6} & 89.5$\pm$0.6          & 89.5$\pm$0.6          \\
EuroSAT       & 86.1$\pm$0.6          & 86.1$\pm$0.6 & 86.3$\pm$0.6          & 89.1$\pm$0.6          & \textBF{89.2$\pm$0.6} & 89.0$\pm$0.6          \\
ISIC          & 48.4$\pm$0.9          & 48.3$\pm$0.9 & \textBF{48.9$\pm$0.9} & 48.6$\pm$0.9          & 48.5$\pm$0.9          & \textBF{48.9$\pm$0.9} \\
ChestX        & \textBF{27.2$\pm$0.6} & 27.1$\pm$0.6 & 27.1$\pm$0.6          & 26.7$\pm$0.6          & 26.6$\pm$0.6          & 26.8$\pm$0.6          \\
Food101       & 53.9$\pm$1.1          & 53.9$\pm$1.1 & 53.9$\pm$1.1          & 61.7$\pm$1.0          & \textBF{61.8$\pm$1.0} & 61.6$\pm$1.0          \\ \hline
mean SG acc   & 65.13                & 65.16       & 65.26                & 69.51                & \textBF{69.54}       & 69.43                \\
\bottomrule
\end{tabular}
\end{table}

\begin{table}[t]
\centering
\caption{Comparison between applying FES to an ImageNet-pretrained EfficientNetV2 extractor alone and applying FES to an extractor collection containing it and the seven other ResNet18 source domain extractors.}
\label{table:efficientnetv2_only}
\begin{tabular}{lcccccc}
\toprule
              & \multicolumn{3}{c}{EfficientNetV2 only}                    & \multicolumn{3}{c}{EfficientNetV2 and ResNet18s}  \\
Dataset       & FES                  & ConFES               & ReFES                & FES                  & ConFES               & ReFES       \\ \hline
ilsvrc\_2012  & \textBF{64.0$\pm$1.0} & \textBF{64.0$\pm$1.0} & 63.9$\pm$1.0          & 63.5$\pm$1.0          & 63.7$\pm$1.0          & 63.1$\pm$1.1          \\
omniglot      & 58.0$\pm$1.3          & 58.1$\pm$1.3          & 57.6$\pm$1.3          & 93.0$\pm$0.7          & \textBF{93.5$\pm$0.6} & 93.3$\pm$0.6          \\
aircraft      & 63.9$\pm$1.0          & 63.8$\pm$1.0          & 63.6$\pm$1.0          & 87.4$\pm$0.8          & \textBF{87.7$\pm$0.7} & 87.5$\pm$0.8          \\
cu\_birds     & 76.0$\pm$0.8          & 76.0$\pm$0.8          & 76.0$\pm$0.8          & 80.6$\pm$0.8          & \textBF{80.9$\pm$0.8} & 80.4$\pm$0.9          \\
dtd           & \textBF{82.2$\pm$0.6} & \textBF{82.2$\pm$0.6} & 82.1$\pm$0.6          & 81.5$\pm$0.8          & 81.7$\pm$0.7          & 81.0$\pm$0.8          \\
quickdraw     & 60.0$\pm$1.0          & 60.0$\pm$1.0          & 59.8$\pm$1.0          & \textBF{83.3$\pm$0.6} & \textBF{83.3$\pm$0.6} & 83.2$\pm$0.6          \\
fungi         & 51.0$\pm$1.2          & 51.0$\pm$1.2          & 50.9$\pm$1.2          & 70.0$\pm$1.1          & \textBF{70.2$\pm$1.1} & 70.0$\pm$1.1          \\
vgg\_flower   & \textBF{97.2$\pm$0.2} & \textBF{97.2$\pm$0.2} & \textBF{97.2$\pm$0.2} & 96.2$\pm$0.5          & 96.9$\pm$0.3          & 96.2$\pm$0.5          \\ \hline
mean WG acc   & 69.04                & 69.04                & 68.89                & 81.94                & \textBF{82.24}       & 81.84                \\ \hline
traffic\_sign & 60.3$\pm$1.2          & 60.2$\pm$1.2          & 60.5$\pm$1.2          & \textBF{65.6$\pm$1.1} & 65.5$\pm$1.1          & 65.4$\pm$1.2          \\
mscoco        & 61.1$\pm$1.0          & \textBF{61.2$\pm$1.0} & 60.6$\pm$1.0          & 60.7$\pm$1.0          & 60.8$\pm$1.0          & 60.0$\pm$1.0          \\
mnist         & 87.1$\pm$0.7          & 87.1$\pm$0.7          & 87.2$\pm$0.7          & 96.3$\pm$0.5          & 96.5$\pm$0.5          & \textBF{96.6$\pm$0.5} \\
cifar10       & \textBF{83.3$\pm$0.6} & \textBF{83.3$\pm$0.6} & 83.2$\pm$0.6          & 82.9$\pm$0.7          & 83.0$\pm$0.8          & 82.7$\pm$0.8          \\
cifar100      & 73.7$\pm$0.9          & 73.7$\pm$0.9          & 73.7$\pm$0.9          & 73.9$\pm$1.0          & \textBF{74.0$\pm$1.0} & 73.8$\pm$1.0          \\
CropDisease   & 85.5$\pm$0.7          & 85.4$\pm$0.7          & 85.4$\pm$0.7          & \textBF{89.6$\pm$0.6} & 89.5$\pm$0.6          & 89.5$\pm$0.6          \\
EuroSAT       & 87.1$\pm$0.6          & 87.0$\pm$0.6          & 87.0$\pm$0.6          & 89.1$\pm$0.6          & \textBF{89.2$\pm$0.6} & 89.0$\pm$0.6          \\
ISIC          & 46.1$\pm$0.9          & 46.0$\pm$0.9          & 46.0$\pm$0.9          & 48.6$\pm$0.9          & 48.5$\pm$0.9          & \textBF{48.9$\pm$0.9} \\
ChestX        & 25.2$\pm$0.5          & 25.2$\pm$0.5          & 25.1$\pm$0.5          & 26.7$\pm$0.6          & 26.6$\pm$0.6          & \textBF{26.8$\pm$0.6} \\
Food101       & 61.2$\pm$1.0          & 61.2$\pm$1.0          & 61.0$\pm$1.0          & 61.7$\pm$1.0          & \textBF{61.8$\pm$1.0} & 61.6$\pm$1.0          \\ \hline
mean SG acc   & 67.06                & 67.03                & 66.97                & 69.51                & \textBF{69.54}       & 69.43                \\
\bottomrule
\end{tabular}
\end{table}

FES and its variants operate in logit space, which means they are independent of the architecture and feature size of each extractor. Therefore, they can naturally work with heterogeneous extractor collections. We demonstrate this by replacing the ResNet18 ImageNet extractor in the source domain collection with a more advanced Small EfficientNetV2 model~\citep{tan2021efficientnetv2} pretrained on the 21K-class version of ImageNet, while keeping the seven other source domain ResNet18 extractors unchanged. The Small EfficientNetV2 model produces feature vectors of length 1280, as opposed to feature vectors of length 512 generated by ResNet18.

URL-style fine-tuning is used, i.e., a square matrix is used for feature projection and the matrix is initialised as an identity matrix. The results are shown in Table \ref{table:efficientnetv2}, and are compared to results of all eight extractors being ResNet18 models. Usage of the EfficientNetV2 model consistently improves FES performance in both weak and strong generalisation. Note that the evaluation's main purpose is to show FES compatibility with heterogeneous model zoos, and its results are not directly comparable to the main results because the 21K-class ImageNet dataset used to pretrain the EfficientNetV2 model contains the Meta-Dataset ImageNet test split, which makes the ImageNet evaluation over-optimistic; moreover, test classes in the other domains may also be present in the 21K pretraining classes.

Since the EfficientNetV2 model is much more advanced than ResNet18, we investigate whether it dominates the extractor collection and effectively makes the other ResNet18 extractors irrelevant by performing FES using the single EfficientNetV2 extractor, with results in Table \ref{table:efficientnetv2_only}. Interestingly, all three FES variants obtain very similar accuracy when applied to only one EfficientNetV2 extractor, while their differences are shown more clearly when applied to a collection of eight extractors. Although using only EfficientNetV2 leads to better performance in a number of ImageNet-adjacent domains, e.g., ilsvrc\_2012, dtd, vgg\_flower, mscoco, and cifar10, it under-performs in most other domains, especially those significantly different from ImageNet, e.g., omniglot, aircraft, quickdraw, fungi, traffic\_sign, mnist, CropDisease, ISIC, and ChestX.

Our EfficientNetV2 evaluation indicates: 1) FES and its variants are compatible with heterogeneous extractor collections, and 2) they are robust to discrepancies in extractor architectures and able to select relevant models from a diverse model zoo.

\section{Limitations and discussion}

\begin{table}[t]
\centering
\caption{ResNet152 feature extractors with URL fine-tuning.}
\label{table:resnet152_url}
\begin{tabular}{lc@{\hspace{0.2cm}}c@{\hspace{0.2cm}}c@{\hspace{0.2cm}}c@{\hspace{0.2cm}}c@{\hspace{0.2cm}}c}
\toprule
Dataset       & URL\_18              & MDL\_152             & URL\_152    & FES\_18              & ConFES\_18           & ReFES\_18            \\ \hline
ilsvrc\_2012  & 56.6$\pm$1.1 & \textBF{59.3$\pm$1.1} & \textBF{59.3$\pm$1.1} & 56.1$\pm$1.1          & 56.1$\pm$1.1          & 55.9$\pm$1.2          \\
omniglot      & 94.5$\pm$0.4 & 94.5$\pm$0.4          & \textBF{94.8$\pm$0.4} & 93.1$\pm$0.6          & 93.4$\pm$0.6          & 93.5$\pm$0.6          \\
aircraft      & 87.7$\pm$0.5 & 90.7$\pm$0.4          & \textBF{91.3$\pm$0.4} & 87.1$\pm$0.8          & 87.5$\pm$0.7          & 87.5$\pm$0.7          \\
cu\_birds     & 80.7$\pm$0.7 & \textBF{85.0$\pm$0.6} & 84.9$\pm$0.6          & 79.1$\pm$0.8          & 79.2$\pm$0.8          & 79.0$\pm$0.8          \\
dtd           & 76.1$\pm$0.6 & 77.7$\pm$0.6          & \textBF{78.3$\pm$0.6} & 75.0$\pm$0.8          & 75.1$\pm$0.8          & 74.9$\pm$0.8          \\
quickdraw     & 82.0$\pm$0.6 & \textBF{83.3$\pm$0.6} & \textBF{83.3$\pm$0.6} & 83.2$\pm$0.6          & 83.2$\pm$0.6          & 83.1$\pm$0.6          \\
fungi         & 69.5$\pm$1.1 & 73.3$\pm$1.0          & \textBF{74.9$\pm$1.0} & 69.6$\pm$1.1          & 69.8$\pm$1.1          & 69.6$\pm$1.1          \\
vgg\_flower   & 91.4$\pm$0.5 & \textBF{93.3$\pm$0.4} & 91.9$\pm$0.5          & 90.7$\pm$0.7          & 90.5$\pm$0.7          & 90.8$\pm$0.6          \\ \hline
mean WG acc   & 79.81       & 82.14                & \textBF{82.34}       & 79.24                & 79.35                & 79.29                \\ \hline
traffic\_sign & 62.6$\pm$1.2 & 57.2$\pm$1.2          & 61.9$\pm$1.2          & 65.2$\pm$1.2          & \textBF{65.3$\pm$1.2} & 65.0$\pm$1.2          \\
mscoco        & 52.7$\pm$1.0 & 52.3$\pm$1.0          & \textBF{55.1$\pm$1.0} & 52.6$\pm$1.0          & 52.7$\pm$1.0          & 52.8$\pm$1.0          \\
mnist         & 94.6$\pm$0.4 & 93.6$\pm$0.5          & 92.9$\pm$0.5          & 96.3$\pm$0.5          & 96.4$\pm$0.5          & \textBF{96.5$\pm$0.5} \\
cifar10       & 71.4$\pm$0.8 & 73.2$\pm$0.7          & \textBF{75.7$\pm$0.7} & 71.7$\pm$0.8          & 71.9$\pm$0.8          & 71.9$\pm$0.8          \\
cifar100      & 62.6$\pm$1.1 & 65.7$\pm$1.0          & \textBF{67.3$\pm$1.0} & 62.7$\pm$1.1          & 62.7$\pm$1.1          & 62.8$\pm$1.1          \\
CropDisease   & 80.5$\pm$0.8 & 81.0$\pm$0.8          & 82.0$\pm$0.7          & 87.2$\pm$0.7          & 87.2$\pm$0.7          & \textBF{87.4$\pm$0.7} \\
EuroSAT       & 86.6$\pm$0.5 & 86.1$\pm$0.6          & \textBF{87.0$\pm$0.5} & 86.1$\pm$0.6          & 86.1$\pm$0.6          & 86.3$\pm$0.6          \\
ISIC          & 45.5$\pm$0.8 & 45.5$\pm$0.8          & 45.4$\pm$0.8          & 48.4$\pm$0.9          & 48.3$\pm$0.9          & \textBF{48.9$\pm$0.9} \\
ChestX        & 26.6$\pm$0.6 & 25.9$\pm$0.5          & 26.3$\pm$0.5          & \textBF{27.2$\pm$0.6} & 27.1$\pm$0.6          & 27.1$\pm$0.6          \\
Food101       & 51.9$\pm$1.1 & 55.3$\pm$1.1          & \textBF{55.4$\pm$1.0} & 53.9$\pm$1.1          & 53.9$\pm$1.1          & 53.9$\pm$1.1          \\ \hline
mean SG acc   & 63.50       & 63.58                & 64.90                & 65.13                & 65.16                & \textBF{65.26}       \\
\bottomrule
\end{tabular}
\end{table}

\begin{table}[t]
\centering
\caption{ResNet152 feature extractors with TSA fine-tuning. Note that the ``URL" in the column titles refers to the universal representation learning process used to meta-train the models, not the feature projection fine-tuning used by the URL algorithm during meta-test. All methods in this table use TSA fine-tuning.}
\label{table:resnet152_tsa}
\begin{tabular}{lc@{\hspace{0.2cm}}c@{\hspace{0.2cm}}c@{\hspace{0.2cm}}c@{\hspace{0.2cm}}c@{\hspace{0.2cm}}c}
\toprule
Dataset       & URL\_18              & MDL\_152             & URL\_152    & FES\_18               & ConFES\_18            & ReFES\_18    \\ \hline
ilsvrc\_2012  & 56.8$\pm$1.1          & 59.3$\pm$1.1          & \textBF{59.9$\pm$1.1} & 56.2$\pm$1.1          & 56.3$\pm$1.1          & 56.2$\pm$1.2          \\
omniglot      & \textBF{95.0$\pm$0.4} & 94.7$\pm$0.4          & \textBF{95.0$\pm$0.4} & 93.3$\pm$0.6          & 93.3$\pm$0.7          & 93.6$\pm$0.6          \\
aircraft      & 88.4$\pm$0.5          & 91.2$\pm$0.4          & \textBF{92.2$\pm$0.4} & 87.6$\pm$0.8          & 87.9$\pm$0.7          & 87.9$\pm$0.8          \\
cu\_birds     & 81.5$\pm$0.7          & 84.8$\pm$0.6          & \textBF{85.0$\pm$0.6} & 79.9$\pm$0.8          & 80.0$\pm$0.8          & 79.8$\pm$0.9          \\
dtd           & 77.1$\pm$0.7          & 78.9$\pm$0.7          & \textBF{79.2$\pm$0.7} & 76.2$\pm$0.8          & 76.3$\pm$0.8          & 76.4$\pm$0.8          \\
quickdraw     & 82.0$\pm$0.6          & 83.4$\pm$0.6          & 83.4$\pm$0.6          & 83.4$\pm$0.6          & \textBF{83.5$\pm$0.6} & 83.4$\pm$0.6          \\
fungi         & 68.3$\pm$1.1          & 73.0$\pm$1.0          & \textBF{74.5$\pm$1.0} & 69.4$\pm$1.1          & 69.7$\pm$1.1          & 69.6$\pm$1.1          \\
vgg\_flower   & 92.1$\pm$0.5          & \textBF{93.4$\pm$0.5} & 92.2$\pm$0.6          & 91.9$\pm$0.7          & 91.9$\pm$0.7          & 92.1$\pm$0.6          \\ \hline
mean WG acc   & 80.15                & 82.34                & \textBF{82.68}       & 79.74                & 79.86                & 79.88                \\ \hline
traffic\_sign & 82.8$\pm$0.9          & 76.4$\pm$1.0          & 80.0$\pm$0.9          & 84.9$\pm$1.0          & 85.1$\pm$1.0          & \textBF{85.2$\pm$1.0} \\
mscoco        & 53.8$\pm$1.1          & 54.1$\pm$1.0          & \textBF{56.9$\pm$1.0} & 54.1$\pm$1.0          & 54.4$\pm$1.0          & 54.4$\pm$1.0          \\
mnist         & 96.6$\pm$0.4          & 95.5$\pm$0.5          & 95.4$\pm$0.5          & 97.1$\pm$0.5          & 97.1$\pm$0.5          & \textBF{97.2$\pm$0.5} \\
cifar10       & 79.9$\pm$0.8          & 79.3$\pm$0.8          & \textBF{81.6$\pm$0.7} & 78.1$\pm$0.9          & 78.2$\pm$0.9          & 78.8$\pm$0.9          \\
cifar100      & 70.3$\pm$1.0          & 70.5$\pm$1.0          & \textBF{73.1$\pm$1.0} & 70.4$\pm$1.1          & 70.6$\pm$1.0          & 70.8$\pm$1.0          \\
CropDisease   & 84.4$\pm$0.8          & 82.7$\pm$0.8          & 84.4$\pm$0.7          & 88.1$\pm$0.7          & \textBF{88.3$\pm$0.7} & \textBF{88.3$\pm$0.7} \\
EuroSAT       & \textBF{89.6$\pm$0.5} & 88.6$\pm$0.6          & \textBF{89.6$\pm$0.5} & 88.8$\pm$0.6          & 89.1$\pm$0.6          & 89.3$\pm$0.6          \\
ISIC          & 48.4$\pm$0.9          & 45.6$\pm$0.9          & 47.3$\pm$0.9          & \textBF{49.5$\pm$0.9} & 49.3$\pm$0.9          & 48.9$\pm$0.9          \\
ChestX        & 27.2$\pm$0.6          & 25.5$\pm$0.6          & 27.3$\pm$0.6          & 27.7$\pm$0.6          & 27.7$\pm$0.6          & \textBF{27.8$\pm$0.6} \\
Food101       & 53.4$\pm$1.2          & 56.2$\pm$1.1          & \textBF{57.3$\pm$1.1} & 55.2$\pm$1.1          & 55.5$\pm$1.1          & 55.2$\pm$1.1          \\ \hline
mean SG acc   & 68.64                & 67.44                & 69.29                & 69.39                & 69.53                & \textBF{69.59}       \\
\bottomrule
\end{tabular}
\end{table}

\begin{table}[t]
\centering
\caption{Computational resource consumption of FES variants using TSA fine-tuning, compared to the official TSA algorithm applied to a URL ResNet18 or ResNet152 extractor. From left to right: CDFSL method, fine-tuning time, stacking classifier training time, number of pretrained parameters that are frozen during fine-tuning, number of trainable parameters during fine-tuning, number of parameters that need to be stored, number of stacking classifier parameters, and amount of GPU memory required for fine-tuning.}
\label{table:computational_cost}
\begin{tabular}{l@{\hspace{0.1cm}}c@{\hspace{0.1cm}}c@{\hspace{0.1cm}}c@{\hspace{0.1cm}}c@{\hspace{0.1cm}}c@{\hspace{0.1cm}}c@{\hspace{0.1cm}}c}
\toprule
method     & FT time                  & MC time & frozen P                      & trainable P                    & stored P                                                                                          & MC P & FT memory \\ \hline
URL\_18    & 10.13s                   & -       & 11M                           & 1.5M                           & \begin{tabular}[c]{@{}c@{}}11M +\\ 1.5M\end{tabular}                                              & -    & 8.2GB      \\ \hline
URL\_152   & 103.88s                  & -       & 60M                           & 7.3M                           & \begin{tabular}[c]{@{}c@{}}60M +\\ 7.3M\end{tabular}                                              & -    & 32.7GB     \\ \hline
FES\_18    & \multirow{3}{*}{151.29s} & 0.06s   & \multirow{3}{*}{11M$\times$8} & \multirow{3}{*}{1.5M$\times$8} & \multirow{3}{*}{\begin{tabular}[c]{@{}c@{}}11M$\times$8 +\\ 1.5M$\times$8$\times$41\end{tabular}} & 328  & 8.3GB      \\
ConFES\_18 &                          & 0.06s   &                               &                                &                                                                                                   & 144  & 8.3GB      \\
ReFES\_18  &                          & 9.45s   &                               &                                &                                                                                                   & 328  & 8.3GB      \\
\bottomrule
\end{tabular}
\end{table}

\begin{table}[t]
\centering
\caption{Comparing the official URL model to a URL model distilled without favouring ImageNet.}
\label{table:url_equal}
\begin{tabular}{lcc}
\toprule
Dataset       & URL\_official        & URL\_equal           \\ \hline
ilsvrc\_2012  & \textBF{56.6$\pm$1.1} & 52.6$\pm$1.1          \\
omniglot      & 94.5$\pm$0.4          & \textBF{95.0$\pm$0.4} \\
aircraft      & 87.7$\pm$0.5          & \textBF{88.6$\pm$0.5} \\
cu\_birds     & \textBF{80.7$\pm$0.7} & 80.0$\pm$0.7          \\
dtd           & \textBF{76.1$\pm$0.6} & 72.0$\pm$0.7          \\
quickdraw     & 82.0$\pm$0.6          & \textBF{82.1$\pm$0.6} \\
fungi         & \textBF{69.5$\pm$1.1} & 68.4$\pm$1.1          \\
vgg\_flower   & \textBF{91.4$\pm$0.5} & 89.5$\pm$0.6          \\ \hline
mean WG acc   & \textBF{79.81}        & 78.53                 \\ \hline
traffic\_sign & 62.6$\pm$1.2          & \textBF{63.0$\pm$1.2} \\
mscoco        & \textBF{52.7$\pm$1.0} & 47.2$\pm$1.0          \\
mnist         & 94.6$\pm$0.4          & \textBF{95.1$\pm$0.4} \\
cifar10       & \textBF{71.4$\pm$0.8} & 66.5$\pm$0.8          \\
cifar100      & \textBF{62.6$\pm$1.1} & 56.9$\pm$1.1          \\
CropDisease   & \textBF{80.5$\pm$0.8} & 79.9$\pm$0.8          \\
EuroSAT       & \textBF{86.6$\pm$0.5} & 83.5$\pm$0.6          \\
ISIC          & \textBF{45.5$\pm$0.8} & 44.8$\pm$0.8          \\
ChestX        & \textBF{26.6$\pm$0.6} & \textBF{26.6$\pm$0.6} \\
Food101       & \textBF{51.9$\pm$1.1} & 49.0$\pm$1.1          \\ \hline
mean SG acc   & \textBF{63.50}        & 61.25                 \\
\bottomrule
\end{tabular}
\end{table}

FES requires no universal extractor, which means the meta-training phase only requires pretraining a collection of extractors, similar to SUR. The cost for this is reduced to zero if pretrained extractors are readily available. However, FES is more expensive in the meta-testing phase in terms of both computation and storage, as it needs to fine-tune each extractor and save their snapshots instead of utilising a single universal extractor. The good performance of FES could be attributed to its increased capacity, as it maintains individual extractors instead of a single universal extractor. In the context of Meta-Dataset, FES maintains eight extractors, which means $8 \times$ parameters compared to a universal model of the same architecture. Hence, in an additional experiment, we investigate larger universal models with capacities comparable to FES.

Originally, \citet{li2021universal} distill eight ResNet18 extractor into a universal ResNet18 extractor. We distilled a universal ResNet152~\citep{he2016deep} extractor using the same process. ResNet18 has 11M parameters while ResNet152 has 60M. We elected to use the same eight ResNet18 extractors for distillation, because pretraining eight ResNet152 extractors from scratch is prohibitively expensive for us, and this avoids introducing a confounding factor to meta-model evaluation because different base-model architectures may encompass source domain semantics differently. We also pretrained a universal ResNet152 model using ``vanilla" multi-domain learning (MDL), i.e., one feature extractor is pretrained with all eight source domains' data using eight classification heads, one for each domain. Compared to official ResNet18 URL training, we halved the mini-batch size (and doubled the number of iterations) to fit ResNet152 URL or MDL training in the 48GB memory of an NVIDIA A6000 GPU---the most advanced at our disposal. Tables \ref{table:resnet152_url} and \ref{table:resnet152_tsa} show their results with URL or TSA fine-tuning respectively, and compare them to using the official ResNet18 URL model, as well as FES variants with ResNet18 extractor collections. As TSA fine-tuning has high memory consumption, we forwent adaptors in the first and second convolutional blocks (shown to have a small impact on accuracy by~\citet{li2022cross}) to fit the ResNet152 TSA experiments on our NVIDIA A6000 GPU. In both tables, the ResNet152 URL model generally outperforms the ResNet18 URL and ResNet152 MDL models, and it achieves best average weak generalisation accuracy. Its mean strong generalisation accuracy is comparable to that of the FES variants, but individual results show that the methods excel at different tasks: the ResNet152 URL model performs better on mscoco, cifar10, cifar100, EuroSAT, and Food101, while the FES methods perform better on traffic\_sign, mnist, CropDisease, ISIC, and ChestX---it appears that the ResNet152 URL model is better at ImageNet-adjacent tasks, while the FES methods are better at tasks that differ more substantially from ImageNet.

Table \ref{table:computational_cost} compares the cost of FES inference using an NVIDIA A6000 GPU to that of URL ResNet18 and ResNet152 extractors. TSA fine-tuning is used by all methods in this table. It is worth pointing out that due to the few-shot nature of each episode, meta-testing is generally not time consuming. Table \ref{table:computational_cost} represents the approximate upper bound of FES computation cost, because 1) the time presented in the table was measured using the largest traffic\_sign episode in our cached sample, which contains 497 support instances, whereas smaller episodes consume less time, 2) URL and FLUTE fine-tuning are much less time-consuming than TSA, and 3) Section \ref{sec:snapshot_omission} shows that a portion of the snapshots does not in fact need to be computed and stored.

FES requires approximately $2 \times K$ as much backpropagation as a universal extractor fine-tuned once, where $2$ represents one fine-tuning run on the cross-validated support set (performed in two splits) and another on the full support set, and $K$ represents the number of extractors. This is reflected in Table \ref{table:computational_cost} as fine-tuning time for the FES methods is approximately 16 times that of fine-tuning the URL ResNet18 model. Time required to train a FES or ConFES stacking classifier is relatively trivial, while ReFES requires more time to determine its regularisation strength using grid search with cross-validation. FES stores multiple snapshots of each extractor during fine-tuning, but not all model parameters need to be saved. Only weights that are updated during fine-tuning need to be saved in snapshots, as the other unchanged weights can be loaded from the original extractor. Common CDFSL fine-tuning algorithms only update a relatively small set of weights: FLUTE fine-tunes batch normalisation weights, URL fine-tunes a feature projection, and TSA fine-tunes channel projections and a feature projection. Therefore, FES snapshots are normally lightweight. Table \ref{table:computational_cost} shows that FES with TSA fine-tuning needs to store approximately 580M parameters---2.32GB---which can fit in most modern GPUs during inference. As FES can fine-tune its extractors sequentially, its memory requirement is comparable to fine-tuning a single extractor with the same method. On the other hand, FES can easily be parallelised to fine-tune multiple extractors at once, should multiple GPUs be available.

Considering computational effort required for meta training, it is worth noting that even though a universal extractor only needs to be trained once, this training process may take days (for ResNet18) to weeks (for ResNet152) on an NVIDIA A6000 GPU; if an individual extractor is added or updated, training of a universal extractor needs to be performed again.

The official URL model was distilled in a process favouring ImageNet by including as many ImageNet instances as the other seven source domains combined in each mini-batch~\citep{li2021universal}. We distilled an alternative URL model while treating all source domains equally. Their comparison is shown in Table \ref{table:url_equal}. The official model performs better in a majority of domains. This indicates that URL distillation may require external knowledge to focus on the right domains to achieve optimal performance. FES and its variants treat all extractors equally {\em a priori} and determine their task-specific relevance based purely on the support set.

\section{Future work}

FES exhibits good CDFSL performance with multiple source domains. It may be feasible to generalise it to other multi-domain learning problems, e.g., multi-domain transfer learning with a more substantial amount of labelled target domain training data.

The heatmaps show that FES generally assigns significant weights to only a small subset of extractor snapshots, implicitly nullifying a majority of snapshots that it deems irrelevant. Pruning strategies may be applied to FES to explicitly eliminate irrelevant snapshots to reduce computational costs.

FES maintains no prior bias to any source domain extractor, and its posterior bias depends on the support set only. In scenarios where prior knowledge is available regarding source and target domain relations, it may be beneficial to enable the user to apply explicit prior biases to certain source domains. This could be achieved in the form of regularisation, e.g., different regularisation pressures are applied to weights associated with different source domains.

\section{Conclusion}

We present the stacking-based CDFSL method FES and the variants ConFES and ReFES. The FES algorithms create snapshots from fine-tuning independent extractors on the support set, use cross-validation to avoid overfitting from support data reuse, and train a simple stacking classifier to appropriately weight the snapshots. FES, ConFES, and ReFES advance the state-of-the-art on the Meta-Dataset benchmark.

Perhaps more importantly, the FES approaches have some practical advantages in real-world scenarios compared to recent methods based on universal models. FES can work with out-of-the-box heterogeneous extractors. If the extractors are readily available, FES does not require their pretraining data down-stream. Its stacking classifier requires little hyperparameter tuning. FES is also computationally cheaper, unless the number of few-shot learning tasks is very large, e.g., in the thousands, where the total cost of performing FES on all tasks begins to exceed that of training a universal model once. Therefore, to field practitioners who wish to use extractors and fine-tuning algorithms specific to their work, FES is likely more flexible and user-friendly than universal-model methods.

\section{Declarations}

\begin{itemize}
    \item Funding: This research is funded by the Ministry of Business, Innovation and Employment of New Zealand as part of a Smart Ideas project entitled ``User-friendly Deep Learning'', please refer to \url{https://www.mbie.govt.nz/science-and-technology/science-and-innovation/funding-information-and-opportunities/investment-funds/endeavour-fund/}.
    \item Conflict of interest/Competing interests: On behalf of all authors, the corresponding author states that there is no conflict of interest.
    \item Ethics approval: Not applicable
    \item Consent to participate: Not applicable
    \item Consent for publication: Not applicable
    \item Availability of data and materials: All data used can be acquired publicly via \url{https://github.com/google-research/meta-dataset} for the official Meta-Dataset, \url{https://github.com/cambridge-mlg/cnaps} for three additional target domains, \url{https://github.com/IBM/cdfsl-benchmark} for four additional target domains, and \url{https://data.vision.ee.ethz.ch/cvl/datasets_extra/food-101/} for one additional target domain.
    \item Code availability: The implementation and the computational work are done using the Python programming language and the PyTorch deep learning library~\citep{NEURIPS2019_9015}. The code and data files are available via GitHub at \url{https://github.com/HongyuJerryWang/FeatureExtractorStacking}.
    \item Authors' contributions: All authors contributed to the study conception and design. Material preparation and data collection and analysis were performed by Hongyu Wang. The first draft of the manuscript was written by Wang and all authors commented on previous versions of the manuscript. All authors read and approved the final manuscript.
    \item Human and Animal Ethics: Not applicable
\end{itemize}

\bibliography{sn-article}

\appendix

\section{Additional heatmaps}\label{sec:appendix_heatmaps}

\begin{figure}[p]
    \centering
    \includegraphics[width=\textwidth]{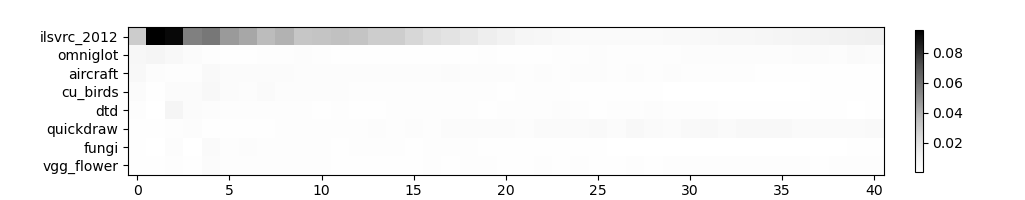}
    \caption{FES kernel for mscoco}
    \label{fig:appendix_start}
    \includegraphics[width=\textwidth]{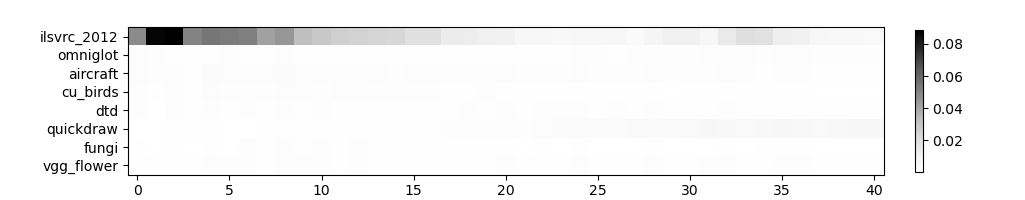}
    \caption{ConFES kernel for mscoco}
    \includegraphics[width=\textwidth]{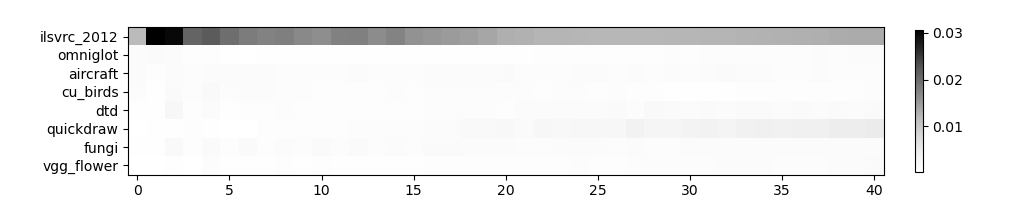}
    \caption{ReFES kernel for mscoco}
\end{figure}

\begin{figure}[t]
    \centering
    \includegraphics[width=\textwidth]{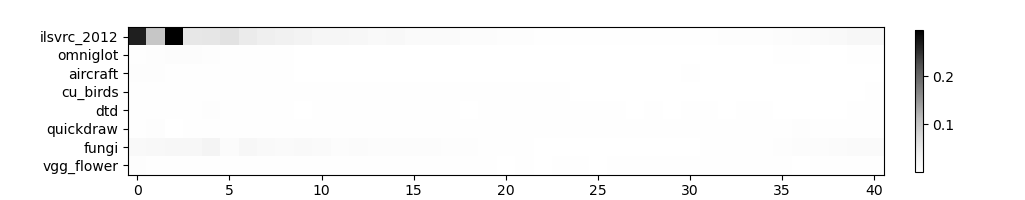}
    \caption{FES kernel for Food101}
    \includegraphics[width=\textwidth]{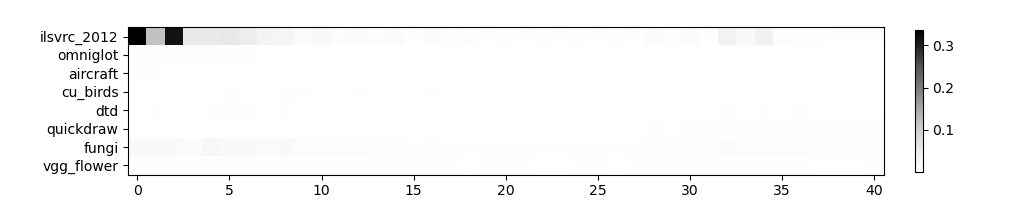}
    \caption{ConFES kernel for Food101}
    \includegraphics[width=\textwidth]{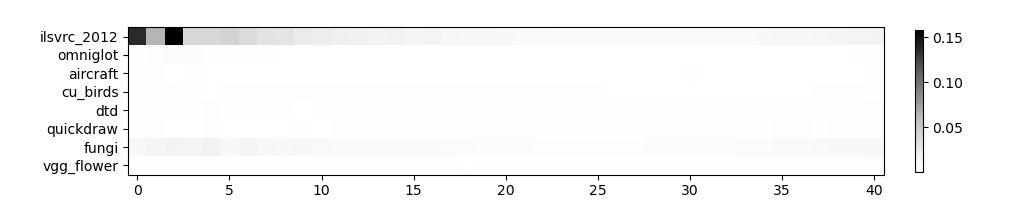}
    \caption{ReFES kernel for Food101}
    \label{fig:appendix_end}
\end{figure}

Additional heatmaps visualising kernel weights on target domains with TSA fine-tuning are shown by Figures \ref{fig:appendix_start}-\ref{fig:appendix_end}.

\begin{figure}[p]
    \includegraphics[width=\textwidth]{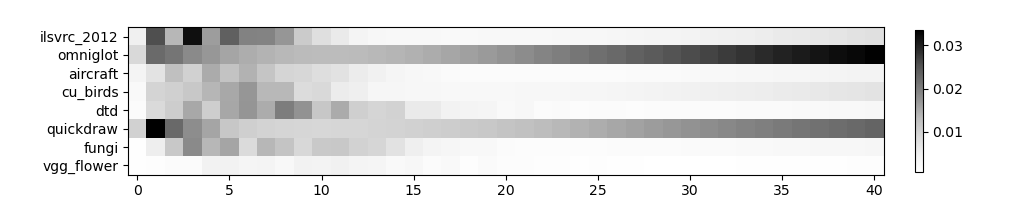}
    \caption{FES kernel for mnist}
    \includegraphics[width=\textwidth]{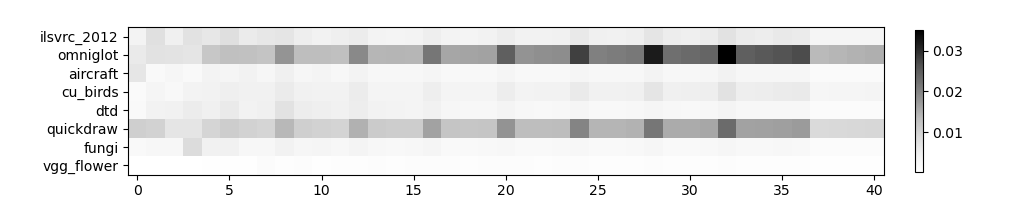}
    \caption{ConFES kernel for mnist}
    \includegraphics[width=\textwidth]{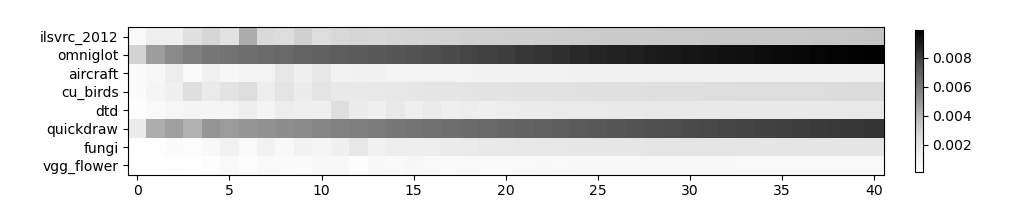}
    \caption{ReFES kernel for mnist}
\end{figure}

\begin{figure}[p]
    \centering
    \includegraphics[width=\textwidth]{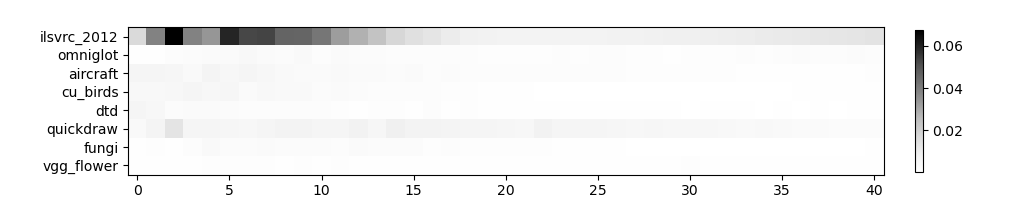}
    \caption{FES kernel for cifar10}
    \includegraphics[width=\textwidth]{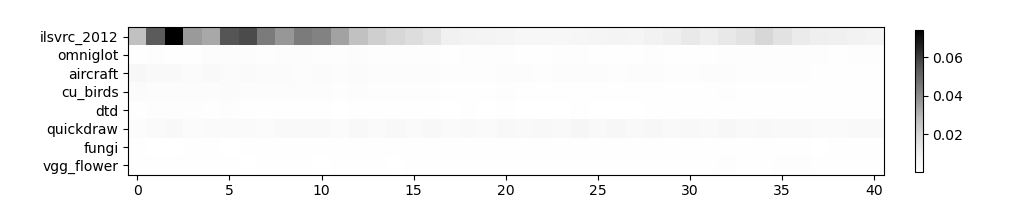}
    \caption{ConFES kernel for cifar10}
    \includegraphics[width=\textwidth]{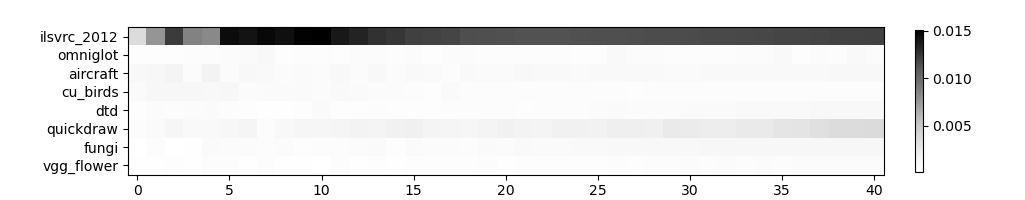}
    \caption{ReFES kernel for cifar10}
\end{figure}

\begin{figure}[p]
    \includegraphics[width=\textwidth]{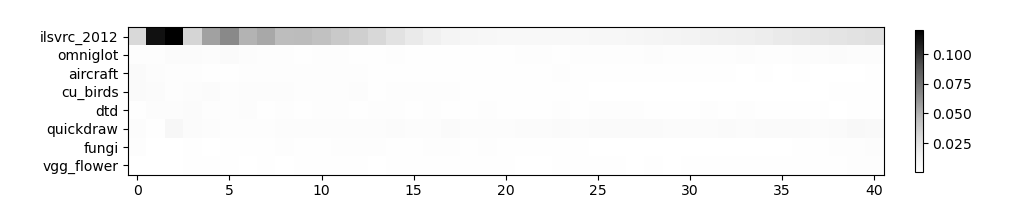}
    \caption{FES kernel for cifar100}
    \includegraphics[width=\textwidth]{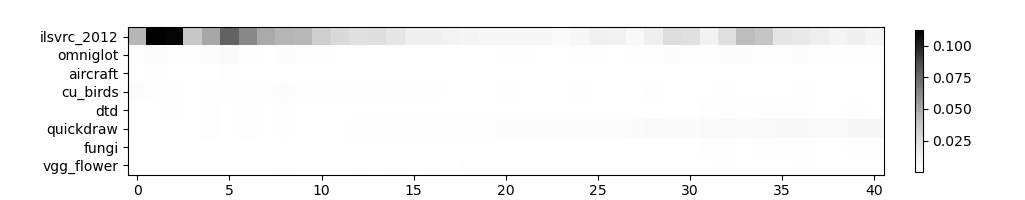}
    \caption{ConFES kernel for cifar100}
    \includegraphics[width=\textwidth]{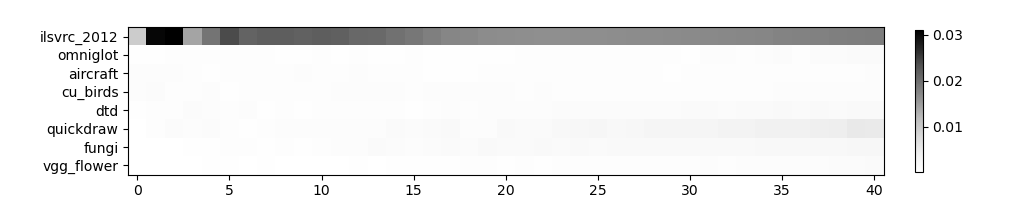}
    \caption{ReFES kernel for cifar100}
\end{figure}

\begin{figure}[p]
    \centering
    \includegraphics[width=\textwidth]{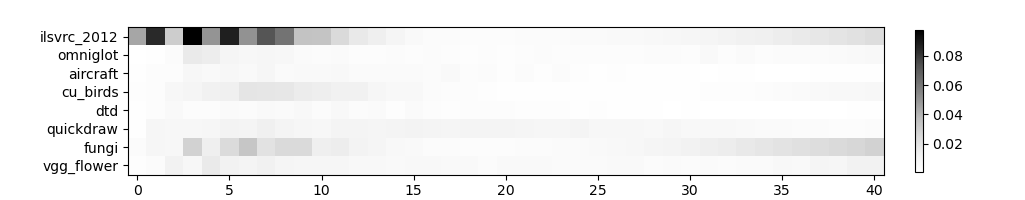}
    \caption{FES kernel for CropDisease}
    \includegraphics[width=\textwidth]{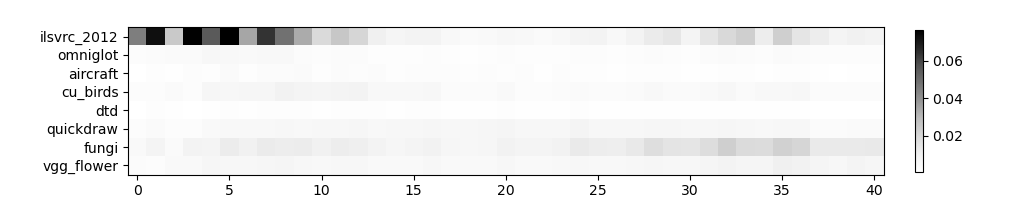}
    \caption{ConFES kernel for CropDisease}
    \includegraphics[width=\textwidth]{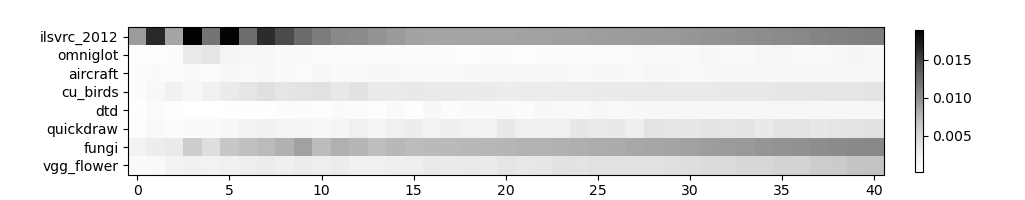}
    \caption{ReFES kernel for CropDisease}
\end{figure}

\begin{figure}[p]
    \includegraphics[width=\textwidth]{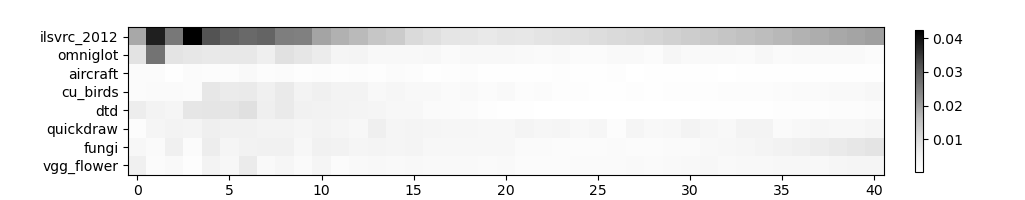}
    \caption{FES kernel for EuroSAT}
    \includegraphics[width=\textwidth]{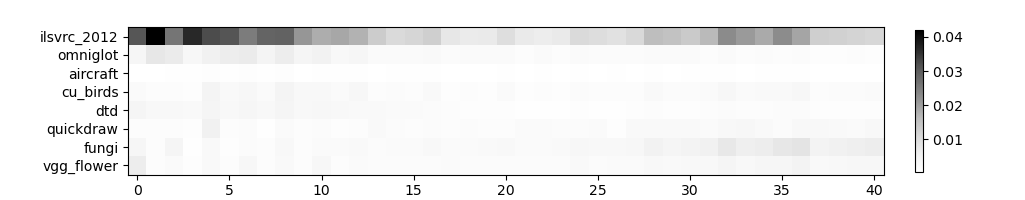}
    \caption{ConFES kernel for EuroSAT}
    \includegraphics[width=\textwidth]{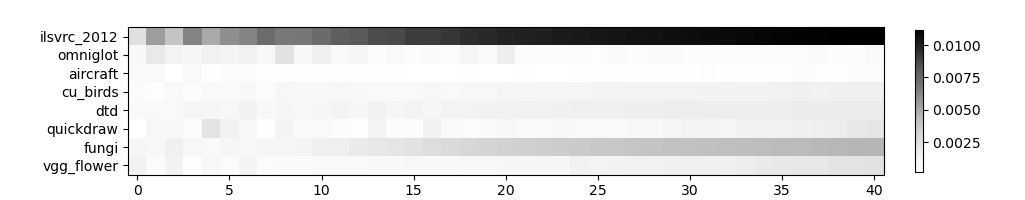}
    \caption{ReFES kernel for EuroSAT}
\end{figure}

\begin{figure}[p]
    \centering
    \includegraphics[width=\textwidth]{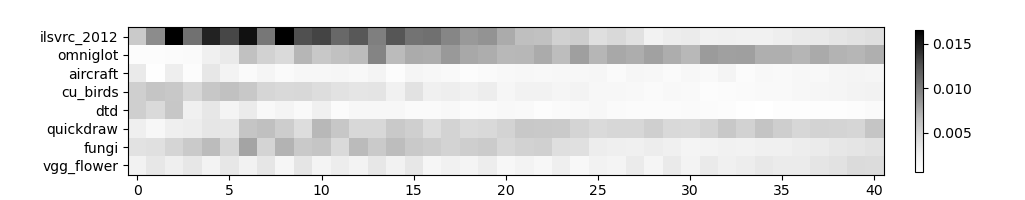}
    \caption{FES kernel for ISIC}
    \includegraphics[width=\textwidth]{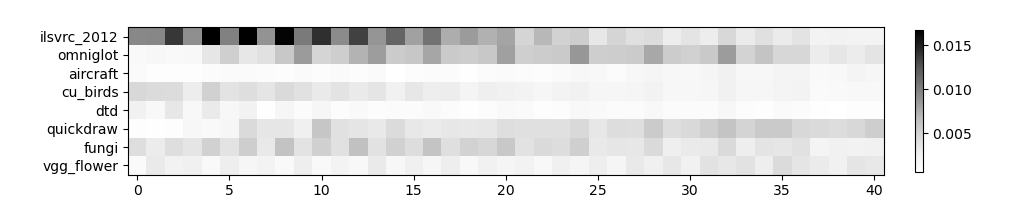}
    \caption{ConFES kernel for ISIC}
    \includegraphics[width=\textwidth]{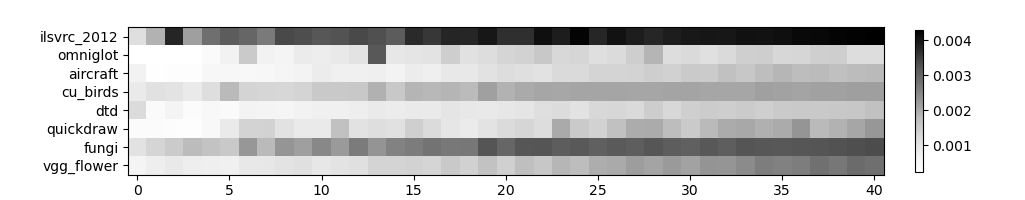}
    \caption{ReFES kernel for ISIC}
\end{figure}

\begin{figure}[p]
    \includegraphics[width=\textwidth]{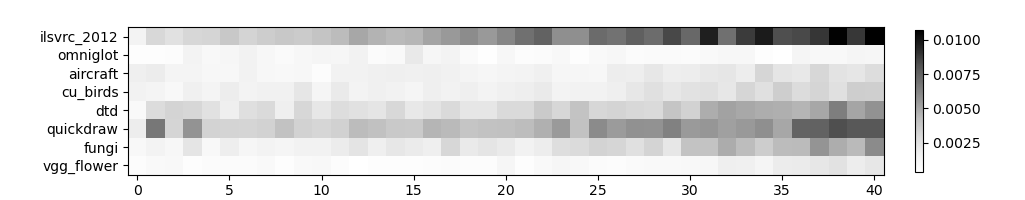}
    \caption{FES kernel for ChestX}
    \includegraphics[width=\textwidth]{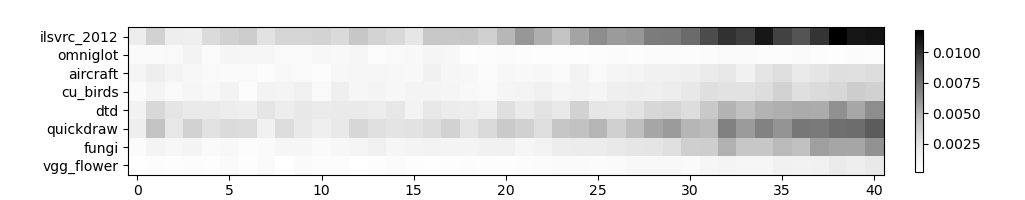}
    \caption{ConFES kernel for ChestX}
    \includegraphics[width=\textwidth]{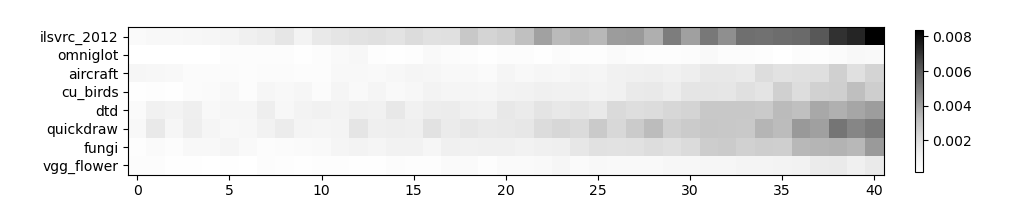}
    \caption{ReFES kernel for ChestX}
\end{figure}

\end{document}